\documentclass[journal]{IEEEtran}
\usepackage{amsmath}
\usepackage{epstopdf}
\usepackage{graphicx}
\usepackage{subfigure}
\ifCLASSINFOpdf
\else
\fi
\begin{document}
%
\title{An Adaptive Parameter Estimation for Guided Filter based Image Deconvolution}
%
%
%

\author{Hang~Yang,
        Zhongbo~Zhang,
        and~Yujing~Guan
\thanks{This work is supported by the National Science Foundation of China under Grant
61401425. Hang Yang is with Changchun Institute of Optics, Fine Mechanics and Physics, Chinese Academy of Science,
Changchun 130033, China. (e-mail:yanghang09@mails.jlu.edu.cn)}
\thanks{Zhongbo Zhang and Yujing Guan are with Jilin University. (e-mail:zhongbozhang@jlu.edu.cn,guanyj@jllu.edu.cn)}}

%
%

\markboth{Journal of \LaTeX\ Class Files,~Vol.~6, No.~1, January~2007}%
{Shell \MakeLowercase{\textit{et al.}}: Bare Demo of IEEEtran.cls for Journals}
%



\maketitle

\begin{abstract}
Image deconvolution is still to be a challenging ill-posed problem for recovering
a clear image from a given blurry image, when the point spread function is known.
Although competitive  deconvolution methods are numerically
impressive and approach theoretical limits, they are becoming more
complex, making analysis, and implementation difficult.
Furthermore, accurate estimation of the regularization parameter is not easy
for successfully solving image deconvolution problems.
In this paper, we develop an effective approach for image
restoration based on one explicit image filter - guided filter.
By applying the decouple of denoising and deblurring techniques to the deconvolution model, we reduce the optimization complexity and achieve a simple but effective algorithm to automatically compute the parameter in
each iteration, which is based on  Morozov's
discrepancy principle. Experimental results demonstrate that
the proposed algorithm outperforms
many state-of-the-art deconvolution methods in terms of both ISNR and visual quality.

\end{abstract}

\begin{IEEEkeywords}
Image deconvolution, guided filter, edge-preserving, adaptive parameter estimation.
\end{IEEEkeywords}

%
\IEEEpeerreviewmaketitle

\section{Introduction}
%
%
%
%
\IEEEPARstart{I}{mage} deconvolution is a classical inverse problem existing in the field of computer vision and image processing\cite{Hansen}. For instance, the image might be captured during the time when the camera is moved, in which case the image is corrupted by motion blur. Image restoration becomes necessary,  which aims at estimating the original
scene from the blurred and noisy observation, and it is one of the most important operations for future processing.

Mathematically, the model of degradation processing is often written as the convolution of the original image with a low-pass filter
\begin{equation}\label{1.1}
    g= \mathcal{H}u_{orig}+ \gamma  =h \ast u_{orig}+\gamma
\end{equation}
where $u_{orig}$ and $g$ are the clear image and the observed image, respectively. $\gamma$ is zero-mean additive white Gaussian noise with variance $\sigma^{2}$. $h$ is the point spread function (PSF) of a linear time-invariant (LTI) system $\mathcal{H}$, and $\ast$ denotes convolution.

The inversion of the blurring is an ill-condition problem, even though the blur kernel $h$ is known, restoring coherent high frequency image details remains be very difficult\cite{Libin}.

It can broadly divide into two classes of image deconvolution methods. The first class comprises of regularized inversion  followed by image denoising, and the second class estimates the desired image based on a variational optimization problem  with some regularity conditions.

The methods of first class apply a regularized inversion of the blur, followed by a denoising approach. In the first step, the
inversion of the blur is performed in Fourier domain. This makes the image sharper, but also has the effect of amplifying the noise, as well as creating correlations in the noise\cite{Schuler}. Then, a powerful denoising method is used to
remove leaked noise and artifacts. Many image denoising
algorithms have been employed for this task: for example, multiresolution transform based method \cite{R.Neelamani,HangY1}, the shape adaptive discrete cosine transform (SA-DCT)\cite{A.Foi}, the Gaussian scale mixture model (GSM) \cite{J.A.Guerrero-Colon}, and the block matching with 3D-filtering(BM3D) \cite{K.Dabov}.

The Total variation (TV) model\cite{ROF,Y.Wang}, $L_{0}$-ABS\cite{J.Portilla}, and SURE-LET\cite{Xue} belong to the second category. The TV model assumes that the $l_{1}$-norm of the  clear image's gradient is small. It is  well known for its edge-preserving property.  Many algorithms based on this model have been proposed in \cite{Y.Wang},\cite{OlegV.Michailovich},\cite{M.Ng}. $L_{0}$-ABS\cite{J.Portilla} is a sparsity-based
deconvolution method exploiting a fixed transform.
SURE-LET\cite{Xue} method uses the minimization of a regularized SURE ($Stein¡¯s$ unbiased risk estimate) for designing deblurring method expressed as a linear expansion of thresholds.  It has also shown that edge-preserving filter based restoration algorithms can also obtain good results in Refs \cite{Hang1, HangJOLT}.

In recent years, the self-similarity and the sparsity of  images are usually integrated to obtain better performance \cite{Dong1,Dong2,Mairal,GaoWen}.
In \cite{Dong1}, the cost function of image decovolution solution incorporates two regularization terms, which separately characterize self-similarity and sparsity, to improve the image quality.
A nonlocally centralized sparse representation (NCSR) method is proposed in \cite{Dong2}, it centralizes the sparse coding coefficients of the blurred image to enhance the performance of image deconvolution.
Low-rank modeling based methods have also
achieved great success in image deconvolution\cite{Dong3,Ji1,Ji2}.
Since the property of image nonlocal self-similarity, similar patches are grouped in a low-rank matrix, then the matrix completion is performed each patch group to recover the image.

The computation of the regularization parameter is another essential issue in our deconvolution process.
By adjusting regularization parameter, a compromise is achieved to preserve the nature of the restored image and suppress the
noise. There are many methods to select the regularization parameter for Tikhonov-regularized problems, however,
most algorithms only fix a predetermined regularization parameter for the whole restoration procsee\cite{ROF},\cite{Osher},\cite{Tai}. 

Nevertheless, in recent works,  a few methods focus on the adaptive
parameter selection for the restoration problem. Usually, the parameter is estimated by the trial-and-error approach,
for instance, the L-curve method \cite{P. C. Hansen},  the majorization-minimization
(MM) approach \cite{M. A.Bioucas}, the generalized cross-validation (GCV) method \cite{Liao}, the variational Bayesian approach \cite{Babacan},  and
 Morozov's discrepancy principle \cite{Wen,Chuan},

In this work \footnote{It is the extension and improvement of the preliminary version presented in
ICIP'13 \cite{Hang1}. It is worth mentioning that the work \cite{Hang1} was identified as falling within the top 10$\%$ of all accepted manuscripts.}, we propose an efficient patch less approach to image
deconvolution problem by exploiting guided filter \cite{He}. Derived from a local linear model, guided filter generates the filtering output by
considering the content of a guidance image. We first integrate this filter into an iterative deconvolution algorithm.
Our method is consisted by two parts: deblurring and denoising. In the denoising step,  two simple cost functions are designed, and
the solutions of them are one noisier estimated image and a less noisy one. The former will be filtered as input image and the latter will work as the guidance image respectively in the denoising step. During the denoising process, the guided filter will be utilized to the output of last step to reduce leaked noise and refine the result of last step.
Furthermore, the regularization parameter plays an important role in our method, and it is essential
for successfully solving ill-posed image deconvolution problems. We apply the Morozov's discrepancy principle to automatically compute regularization parameter at each iteration.
Experiments manifest that our algorithm is effective on finding a good  regularization parameter, and the proposed algorithm outperforms many competitive methods in both ISNR and visual quality.

This paper is organized as follows. A brief overview of the guided filter is given in Section II.
Section III shows how the guided filter is applied to regularize the deconvolution problem and how the regularization parameter is updated.
Simulation results and the effectiveness of choosing regularization parameter are given in Section IV. Finally, a succinct conclusion is drawn in Section V.

\section{Guided Image Filtering}

Recently, the technique for edge-perserving filtering
has been a popular research topic in computer vision and image processing\cite{Shutao}. Edge-aware filters such as bilateral filter \cite{TOMASI}, and L0 smooth filter \cite{XUL} can suppress noise while preserving important structures. The guided filter \cite{He}\cite{He1} is one of the fastest algorithms
for edge-preserving filters, and it is easy to implement.
In this paper, the guided filter is first applied
for image deconvolution.

Now, we introduce guided filter, which involves a filtering input image $u_{p}$,  an filtering output image $u$ and a guidance image $u_{I}$.
Both $u_{I}$ and $u_{p}$ are given beforehand according to the application, and they can be identical.

The key assumption of the guided filter is a local linear model between the
guidance $u_{I}$ and the filtering output $u$. It assumes that $u$ is a linear transform of  $u_{I}$ in a window $\omega_{k}$ centered
at the pixel $k$ (the size of $\omega_{k}$ is $w\times w$.) :
\begin{equation}\label{2.1}
    u(i) = a_{k} u_{I}(i) + b_{k}
\end{equation}
where $(a_{k}, b_{k})$ are some linear coefficients assumed to be constant in $\omega_{k}$.

To compute the linear coefficients $(a_{k}, b_{k})$, it needs constraints from the filtering input $p$. Specifically, one can minimize the following cost function in the window $\omega_{k}$
\begin{equation}\label{2.2}
    E(a_{k}, b_{k}) = \sum_{i \in \omega_{k}} ((a_{k}u_{I}(i)+b_{k}-u_{p}(i))^{2} + \varepsilon a^{2}_{k})
\end{equation}
Here $\varepsilon$ is a regularization parameter penalizing large $a_{k}$.
The solution of Eq.(\ref{2.2}) can be written as:
\begin{eqnarray}
  a_{k} &=& \frac{\frac{1}{w^{2}}\sum_{i \in \omega_{k}}u_{I}(i)u_{p}(i)-\mu_{k}\bar{p}_{k}}{\sigma^{2}_{k}+\varepsilon} \\
  b_{k} &=& \bar{p}_{k} - a_{k}\mu_{k}
\end{eqnarray}
Here, $\mu_{k}$ and $\sigma_{k}$ are the mean and variance of $u_{I}$ in $\omega_{k}$, and $\bar{p}_{k}$ is the mean of $u_{p}$ in $\omega_{k}$. 

However, a pixel $i$ is involved in all the overlapping windows $\omega_{k}$ that cover $i$,
so the value of $u(i)$ in Eq.(\ref{2.1}) is not the same when it is computed in different windows.
A native strategy is to average all the possible values of $u(i)$. So, we compute the filtering output by:
\begin{equation}\label{2.3}
    u(i) = \bar{a}_{i}u_{I}(i) + \bar{b}_{i}
\end{equation}
where $\bar{a}_{i} = \frac{1}{w^{2}}\sum_{k \in \omega_{k}}a_{k} $ and $\bar{b}_{i} = \frac{1}{w^{2}}\sum_{k \in \omega_{k}}b_{k}$ are the average coefficients of all windows overlapping $i$. More details and analysis can be found in \cite{He1}.

We denote Eq.(\ref{2.3}) as $u = \textbf{guidfilter}(u_{I},u_{p})$.


\section{Algorithm}

\begin{figure*}[ht]
  \centering
  \centerline{\includegraphics[width=1\linewidth]{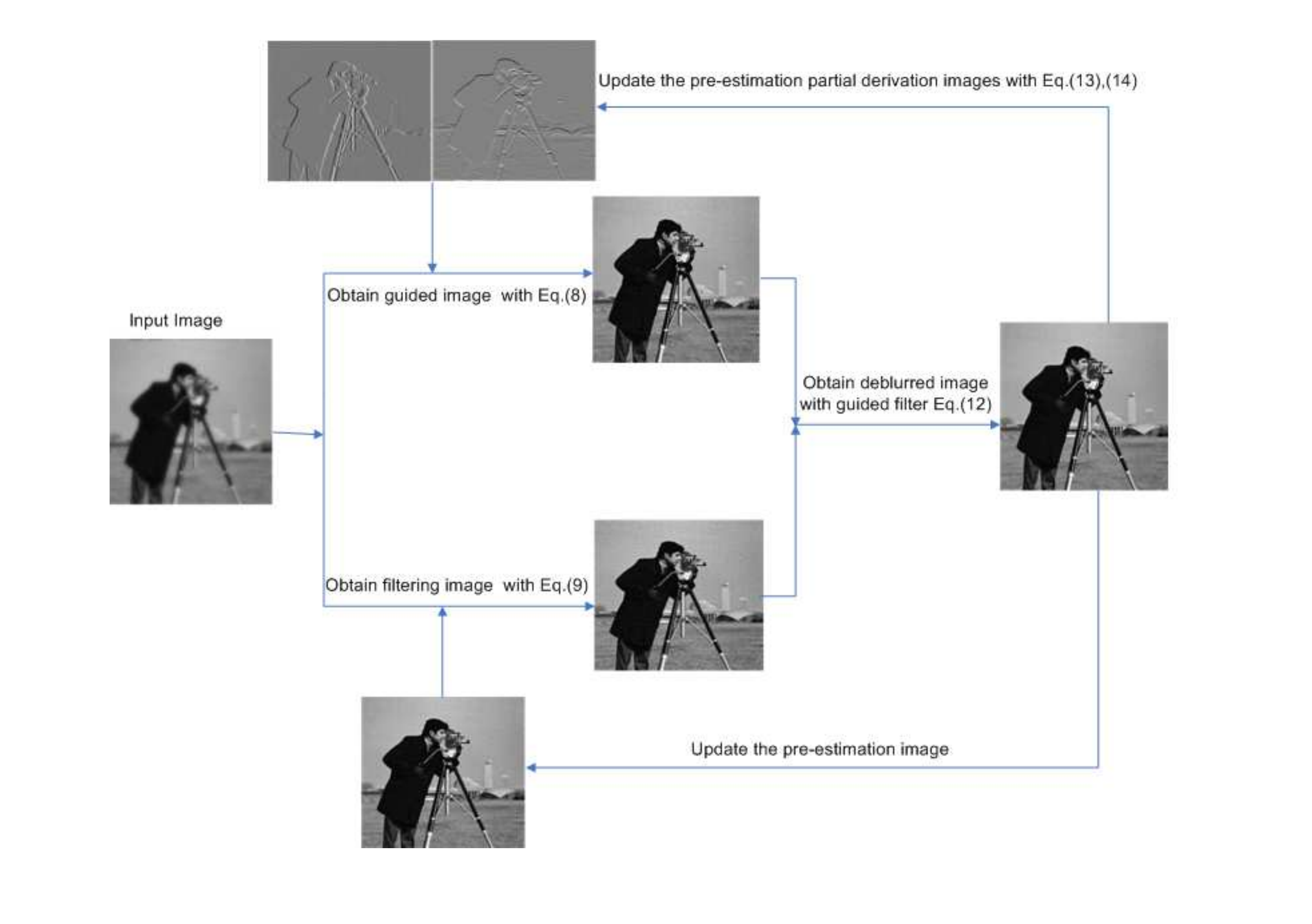}}
\caption{Schematic diagram of the proposed image deconvolution method.}
\label{p.2}
\end{figure*}

\subsection{Guided Image Deconvolution}

In this proposed method, we plan to restore the blurred image by
iteratively Fourier regularization  and image smoothing  via guided filter(GF).
Fig. 1 summarizes the main processes of our
image deconvolution method based on guided filtering (GFD).
The proposed algorithm relies on two steps: (1) two cost functions are proposed to obtain a filtering input image and a guidance image, (2) a sharp image is estimated  using a guided filtering. We will give the detailed description of these two steps in
this section.

In this work, we minimize the following cost function to estimate the original image:
\begin{equation}\label{1.2}
    \min_{u} \parallel g - h \ast u \parallel^{2} + \lambda \parallel u - \textbf{GF}(z, u)\parallel^{2}
\end{equation}
where \textbf{GF} is the guided filter smoothing opteration\cite{He}, and $z$ is the guidance image.
However, directly solving this problem is difficult because $\textbf{GF}(\cdot,\cdot)$ is highly nonlinear.
So, we found that the decouple of deblurring and denoising steps can achieve a good result in practice.

The deblurring step has the positive effect of localizing information, but it has the negative effect of introducing ringing artifacts. In the \textbf{deblurring} step, to obtain a sharper image from the observation, we propose two cost functions:
\begin{eqnarray}
\nonumber
\hspace*{-0.5cm} u_{I} &=& \arg \min_{u} \{ \lambda (\parallel \partial_{x}u-  v_{x} \parallel^{2}_{2} + \parallel \partial_{y}u-  v_{y} \parallel^{2}_{2}) \\
\quad & & + \parallel h \ast u -g  \parallel^{2}_{2}\}\\
\hspace*{-0.5cm}  u_{p} &=& \arg \min_{u}\{ \lambda \parallel u- v \parallel^{2}_{2} + \parallel h \ast u -g  \parallel^{2}_{2}\}
\end{eqnarray}
where $\lambda > 0$ is the regularization parameter, $v$, $v_{x}$ and $v_{y}$ are  pre-estimated natural image,
 partial derivation image in $x$ direction and
  partial derivation image in $y$ direction , respectively. 

Alternatively, we diagonalized derivative operators after Fast Fourier Transform (FFT) for speedup. It can be solved in the Fourier domain
\begin{eqnarray}
\nonumber
  \mathcal{F}(u_{I}) &=& \frac{\mathcal{F}(h)^{*}\cdot \mathcal{F}(g)}{\mid \mathcal{F}(h) \mid^{2}+\lambda(\mid \mathcal{F}(\partial_{x})|^{2} +|\mathcal{F}(\partial_{y})\mid^{2})} +  \\
  & & \lambda \frac{\mathcal{F}(\partial_{x})^{*}\cdot \mathcal{F}(v_{x})+\mathcal{F}(\partial_{y})^{*}\cdot \mathcal{F}(v_{y}) }{\mid \mathcal{F}(h) \mid^{2}+\lambda(\mid \mathcal{F}(\partial_{x})|^{2} +|\mathcal{F}(\partial_{y})\mid^{2})}  \\
  \mathcal{F}(u_{p}) &=& \frac{\mathcal{F}(h)^{*}\cdot \mathcal{F}(g)+\lambda \mathcal{F}(v)}{\mid \mathcal{F}(h) \mid^{2}+\lambda}
\end{eqnarray}
where $\mathcal{F}$ denotes the FFT operator and $\mathcal{F}(\cdot)^{*}$ is the complex conjugate.
The plus, multiplication, and division are all component-wise operators.

 Solving Eq.(11) yields image $u_{p}$ that contains useful high-frequency structures and a special form of distortions. In the alternating minimization process, the high-frequency image details are preserved  in the denoising step, while the noise in $u_{p}$ is gradually reduced.

The goal of \textbf{denoising} is to suppress the amplified noise and artifacts introduced by Eq.(11),  the guided filter is applied to smooth the $u_{p}$, and $u_{I}$ is
used as the guidance image.
Since the guided filter has shown promising performance, it has good edge-aware
smoothing properties near the edges.
Also, it produces distortion free result by
removing the gradient reversals artifacts.
Moreover, the guided filter considers the intensity  information
and texture of neighboring pixels in a window.
In terms of edge-preserving smoothing, it is useful for removing the artifacts. Therefore in our work, the guided filter is integrated to the deconvolution mode
and obtain a reliable sharp image:

\begin{equation}\label{2.5}
    v = \textbf{guidfilter}(u_{I},u_{p})
\end{equation}
The filtering output $v$  is used as the pre-estimation image in Eq.(9).

After the regularized inversion in Fourier domain [see Eq.(10) and (11)], the image $u_{p}$ contains more leaked noise and texture details than $u_{I}$. So we use $u_{p}$ as the filtering input image and $u_{I}$ as the guidance image to reduce the leaked noise and recover some details.

Another problem is how to obtain the pre-estimation image $v_{x}$ and $v_{y}$ in Eq.(8). The first term in Eq.(8) uses image derivatives for reducing
ringing artifacts.  The regularization term $ \parallel \nabla u \parallel$ prefers $u$ with smooth gradients, as mentioned in \cite{Osher}.
A simple method is $v_{x} = \partial_{x}v , v_{y} = \partial_{y}v $ which contains not only high-contrast edges but also enhanced noises.
In this paper, we remove the noise by using the guided image filter again.
Then, we compute the $v_{x}$ and $v_{y}$ as follows:
\begin{eqnarray}\label{2.4}
    v_{x} &=& \textbf{guidfilter}(\partial_{x}v,\partial_{x}v)\\
    v_{y} &=& \textbf{guidfilter}(\partial_{y}v,\partial_{y}v)
\end{eqnarray}
The results can perserve most of the useful information (edges) and ensure spatial consistency of the images, meanwhile suppress the enhanced noise.

The guided filter output is a locally linear transform of the guidance image.
This filter has the edge-preserving smoothing property like the bilateral filter, but does not suffer from the gradient reversal artifacts. So we integrate the guided filter into the deconvolution problem, which leads to a powerful method that produces high quality results.

Meanwhile, the guided filter has a fast and non-approximate linear-time operator, whose computational complexity is only depend on the size of image. It has an $O(N^{2})$ time (in the number of pixels $N^{2}$) exact algorithm for both gray-scale and color images\cite{He}.

We summarize the proposed algorithm as follows :

---------------------------------------------------------

$Step~0$: Set $k=0$,  pre-estimated image $v^{k}=v^{k}_{x}=v^{k}_{y}=0$.

$Step~1$: Iterate on $k = 1,2,...,iter$

\quad 1.1: Use $v^{k}$, $v^{k}_{x}$ and $v^{k}_{y}$ to obtain the filtering input image $u^{k}_{p}$ and the guidance image $u^{k}_{I}$ with  Eq.(8) and Eq.(9), respectively.

\quad 1.2: Apply guided filter to $u^{k}_{p}$ with the guidance image $u^{k}_{I}$ with  Eq.(12), and obtain a filtered output $v^{k+1} = \textbf{guidfilter}(u^{k}_{I},u^{k}_{p})$ .

\quad 1.3: Apply guided filter to $\partial_{x}v^{k}$ and $\partial_{y}v^{k}$ with  Eq.(13) and Eq.(14) to obtain $v^{k}_{x}$ and $v^{k}_{y}$, respectively, and $k = k+1.$

$Step~2$: Output  $v^{iter}$.

-------------------------------------------------------

For initialization, $v^{0}$ is set to be zero (a black image).

\subsection{Choose Regularization Parameter}

Note that the Fourier-based regularized inverse operator in Eq.(12) and (13), and the
deblurred images depend greatly on the regularization parameter $\lambda$. In this subsection, we propose a simple but effective method to estimate the parameter automatically. It depends only on the data and automatically computes
the regularization parameter according to the data.

The Morozov's discrepancy
principle  \cite{S. Anzengruber} is useful for the selection of $\lambda$ when the noise variance is available.
Based on this principle, for an image of  $N\times N $ size, a good estimation of the deconvolution problem
should lie in set
\begin{equation}\label{2.2.1}
    S = \{u;\parallel h \ast u -g  \parallel^{2}_{2}\leq c \}
\end{equation}
where $c = \rho N ^{2}\sigma^{2}$, and $\rho$ is a predetermined parameter.

Indeed, it does not exist uniform criterion for estimating $\rho$
and it is still a pendent problem deserving further study.
For Tikhonov-regularized algorithms, one feasible approach for
selecting $\rho$ is the equivalent degrees of freedom (EDF)
method\cite{Chuan}.

Now, we show how to determine the paremeter $\lambda$. By Parseval's theorem and Eq.(11)
\begin{equation}\label{3.7}
\begin{array}{lll}
    \parallel h \ast u_{p} - g  \parallel^{2}_{2}&=&\parallel \frac{\lambda(\mathcal{F}(h)\cdot \mathcal{F}(v)-\mathcal{F}(g))}{|\mathcal{F}(h)|^{2}+\lambda} \parallel^{2}_{2} \\
    &\leq& \parallel h \ast v -g  \parallel^{2}_{2}
\end{array}
\end{equation}
If the pre-estimated image $v \in S$, then $u_{p} \in S $, so we set $\lambda = \infty $, $u_{I} = u_{p} = v$;
else, a proper parameter $\lambda$ can be computed as follows:  in the case when the additive noise variance is available, a proper regularization parameter $\lambda$ is computed such that the restored image $u_{p}$ in Eq.(11) satisfies
\begin{equation}\label{3.7}
    \parallel h \ast u_{p} -g  \parallel^{2}_{2}=\parallel \frac{\lambda(\mathcal{F}(h)\cdot \mathcal{F}(v)-\mathcal{F}(g))}{|\mathcal{F}(h)|^{2}+\lambda} \parallel^{2}_{2}  = \rho N^{2}\sigma^{2}
\end{equation}

One can see that the right-hand side is monotonically increasing function in $\lambda$, hence we can determine the unique solution \textbf{ via bisection}.

In this paper, the noise variance $\sigma^{2}$ is estimated with the wavelet transform based median rule \cite{Mallat}. Once $\sigma^{2}$ is available, $c = \rho N ^{2}\sigma^{2}$ is generally used to compute $c$. $\rho = 1 $ had been
a common choice. From Eq.(\ref{3.7}),  it is clear that  the $\lambda$ increases with the increase of $\rho$.
In practice, we find that a large $\lambda \ (\rho = 1)$ can substantially cut down the noise variance, but it often causes a noisy image with ringing artifacts.
So, we should select a smaller $\lambda \ (\rho < 1)$ which can achieve an edge-aware image with less noise.
Then, our effective filtering approach based on guided filter can be employed in the denoising step.

In our opinion, the parameter $\rho$ should satisfy an important property: the $\rho$ should decrease with the increase of image variance. For instance, a smooth image which contains few high-frequency information can not produce the strong ringing effects with large $\rho$, and a large $\rho$ could substantially suppress the noise. That is to say, the parameter $\rho$ should increase with the decrease of image variance. According to this property, we set $\rho$ as following:
\begin{equation}\label{3.9}
    \rho = \begin{cases}
    s^{2}, &  thresh > \tau \\
    s, &  thresh \leq \tau
    \end{cases}
\end{equation}
\begin{equation}\label{3.10}
    s = 1-\frac{\parallel g-\mu(g) \parallel_{2}^{2}-N^{2}\sigma^{2}}{\parallel g \parallel_{2}^{2}}
\end{equation}
\begin{equation}\label{3.10}
    thresh = \sqrt{\frac{\parallel g-\mu(g) \parallel_{2}^{2}-N^{2}\sigma^{2}}{N^{2}\sigma^{2}\parallel v - \mu(v) \parallel_{2}^{2}}}
\end{equation}
where $\mu(g)$ and $\mu(v)$ denote the mean of $g$ and $v$, respectively.

This method is simple but practical for computing $\lambda$, it slights adjusting the value of $\lambda$ according to the variance of $g$. $thresh$ shows the ratio of the variance of $g$ and the variance of $v$. It means that when $v$ becomes clearer, the variance of $v$ is larger, and the value of $thresh$ is small.
So we should increase the $\lambda$, because the weight of $v$ is large in Eq.(10) and  Eq.(11).

In this work, we suggest setting $\tau = 0.6$ in our experiments and not adjusting the parameters for different types of images.
We find that this parameter choosing approach is robust to the extent towards
the variations of the image size and the image type. There are some similar
strategies in some works dealing with the constrained TV restoration problem \cite{Chuan}.

\section{Numerical Results}

In this section, some simulation results are conducted
to verify the performance of our algorithm for
image deconvolution application.

We use the BSNR (blurred signal-to-noise
ratio) and the ISNR (improvement in signal-to-noise-ratio) to
evaluate the quality of  the observed and the restored images, respectively. They are defined
as follows:
\begin{eqnarray}
  BSNR &=& 10\log_{10}Var(g)/N^{2}\sigma^{2} \\
  ISNR &=& 10\log_{10}(\frac{\parallel u_{orig}-g \parallel_{2}^{2}}{\parallel u_{orig}- \hat{u} \parallel_{2}^{2}})
\end{eqnarray}
where $\hat{u}$ is the corresponding restored image.

Two $256 \times 256$ ($Cameraman$ and $House$) and two 512$\times$512 ($Lena$ and $Man$) images shown in Fig. \ref{p.2} are used
in the experiments.
\begin{figure}[htb]
  \centering
  \centerline{\includegraphics[width=10.0cm]{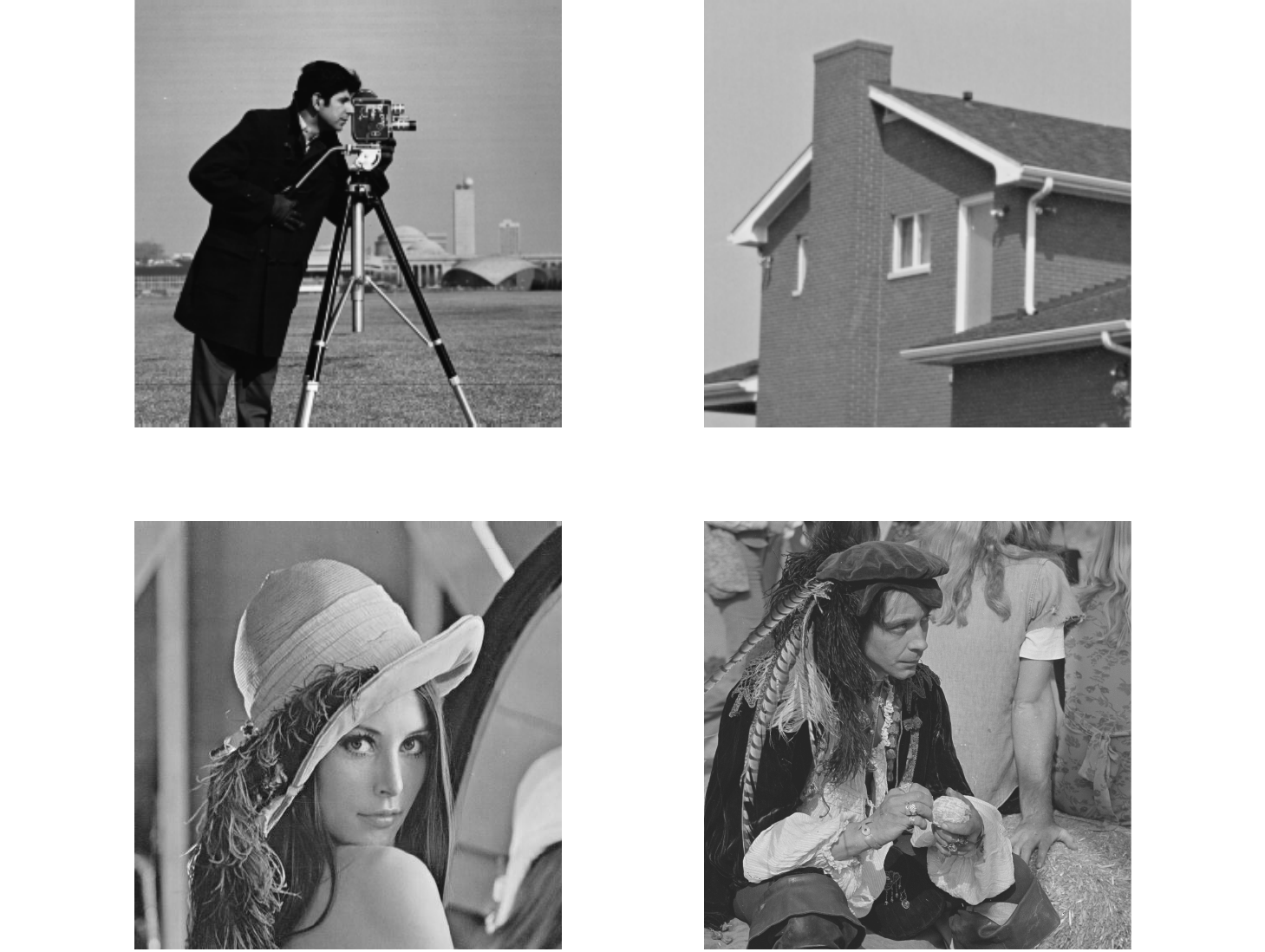}}
\caption{$Cameraman(256 \times 256)$,$House(256 \times 256)$,$Lena(512\times 512)$,$Man(512\times 512)$.}
\label{p.2}
\end{figure}

\subsection{Experiment 1 -- Choice of $ \rho $}

In the first experiment, we explain why  a
suitable upper bound $ c = \rho N^{2}\sigma^{2} $ for the discrepancy
principle is more attractive in image deconvolution. In particular, we demonstrate that the proposed method we derived in Eq.(18) is successful in finding the $\lambda$ automatically.

The test images are $Cameraman$ and $Lena$.
The boxcar blur ($9\times 9$) and the Gaussian blur ($25\times 25$ with std = 1.6) are used in this experiment.
There are many previous methods\cite{J.Portilla}\cite{K.Dabov} for image deconvolution problem have shown the ISNR performance for these two point spread funcionts. We add a Gaussian noise to each blurred image such that the BSNRs of the observed images are 20, 30, and 40 dB, respectively.

The changes of ISNR against $\rho$ is shown in Fig.1. The
ISNRs obtained by  our choice of given in Eq.(18) are
marked by $"\bullet"$ and those obtained by the conventional setting of $ \rho =1$ are marked by
$"\diamondsuit"$. It is clearly shown that, the ISNRs obtained by our approach are always close to the maximum. However, for $\rho =1$,  its ISNRs are far from
the maximum.

\begin{figure*}[ht]
\begin{minipage}[t]{0.3\linewidth}
\centering
\includegraphics[width=1\linewidth]{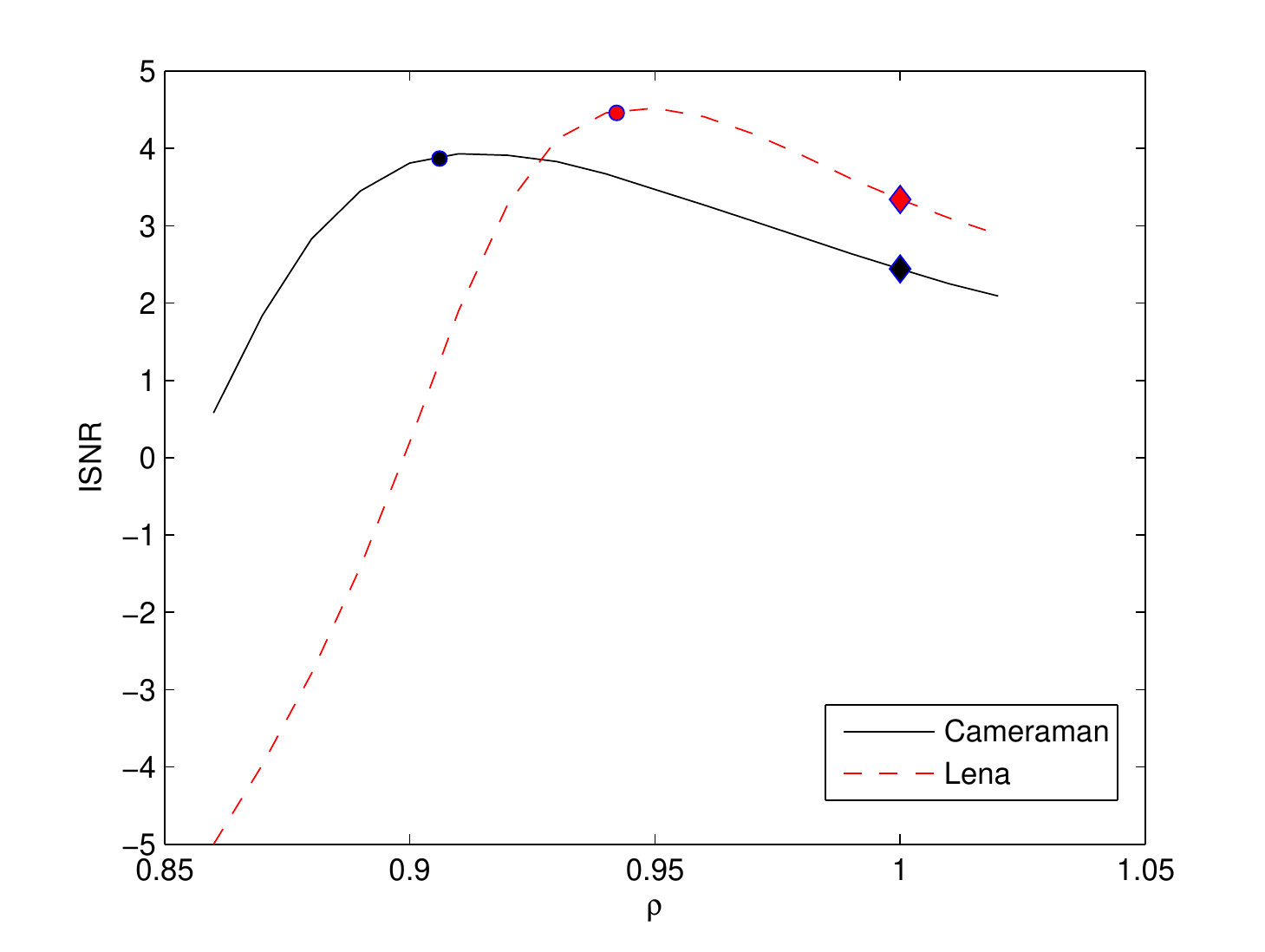}\\
\includegraphics[width=1\linewidth]{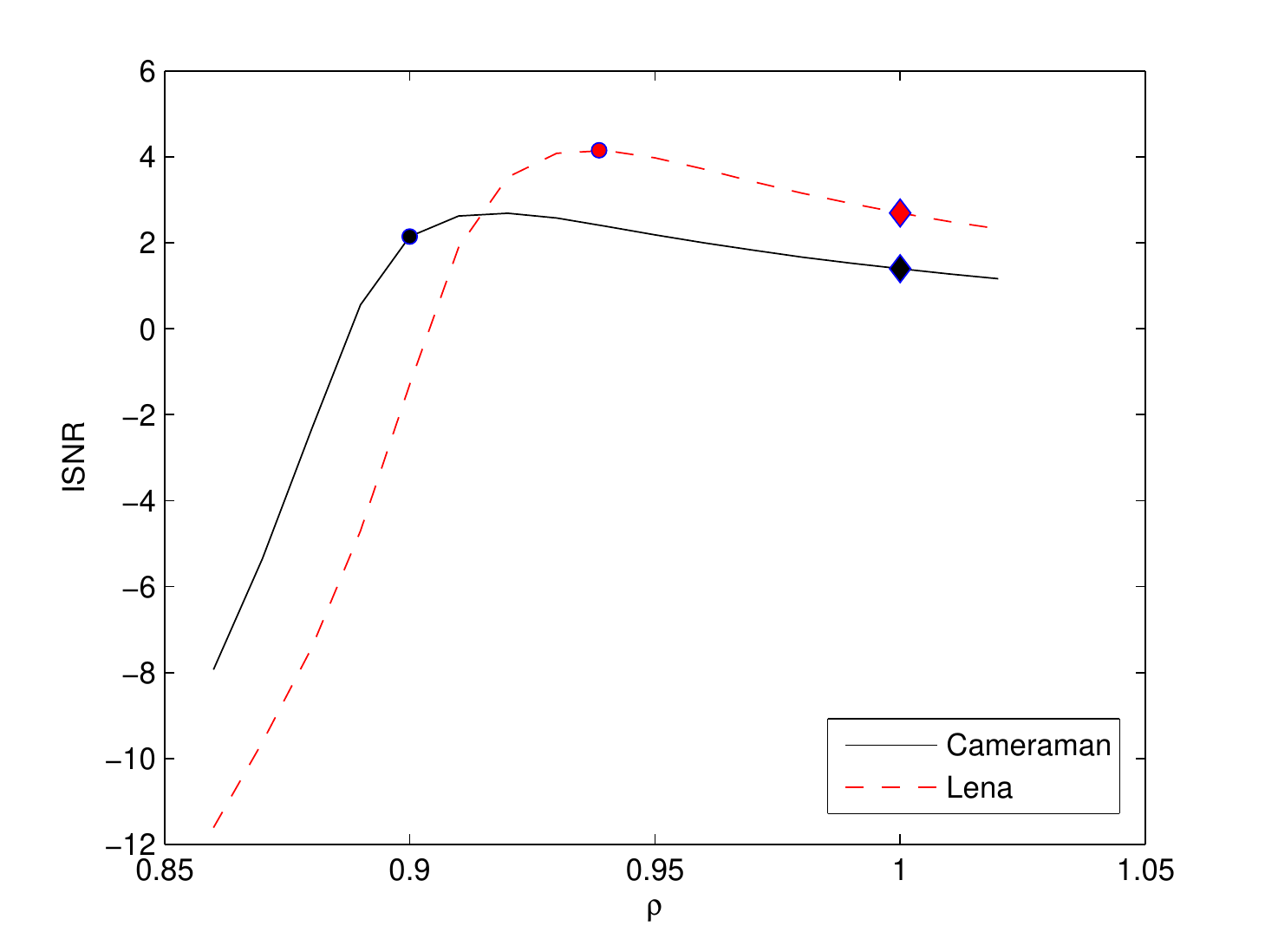}
\label{fig:side:a}
\end{minipage}%
\begin{minipage}[t]{0.3\linewidth}
\centering
\includegraphics[width=1\linewidth]{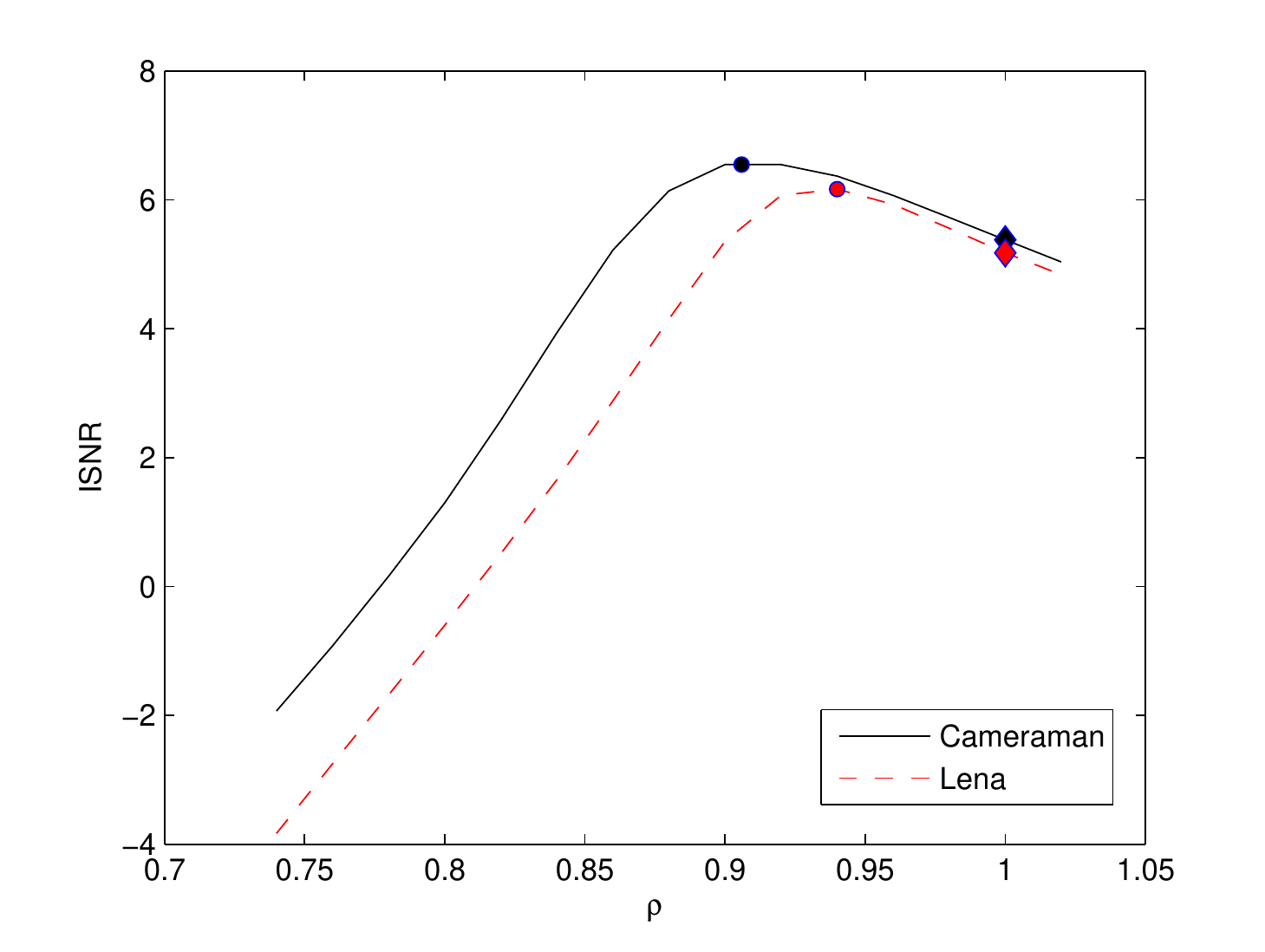}\\
\includegraphics[width=1\linewidth]{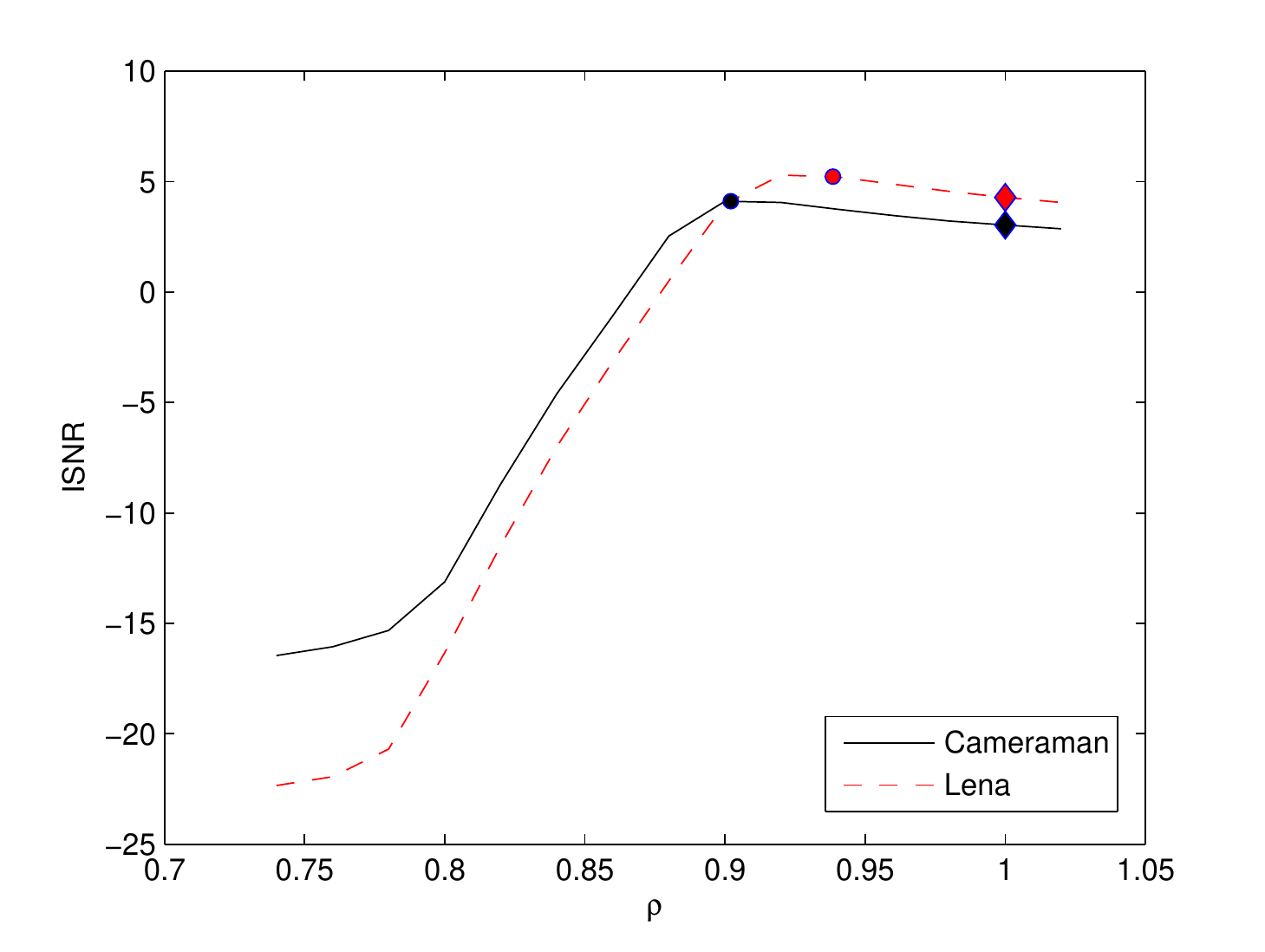}
\label{fig:side:b}
\end{minipage}
\begin{minipage}[t]{0.3\linewidth}
\centering
\includegraphics[width=1\linewidth]{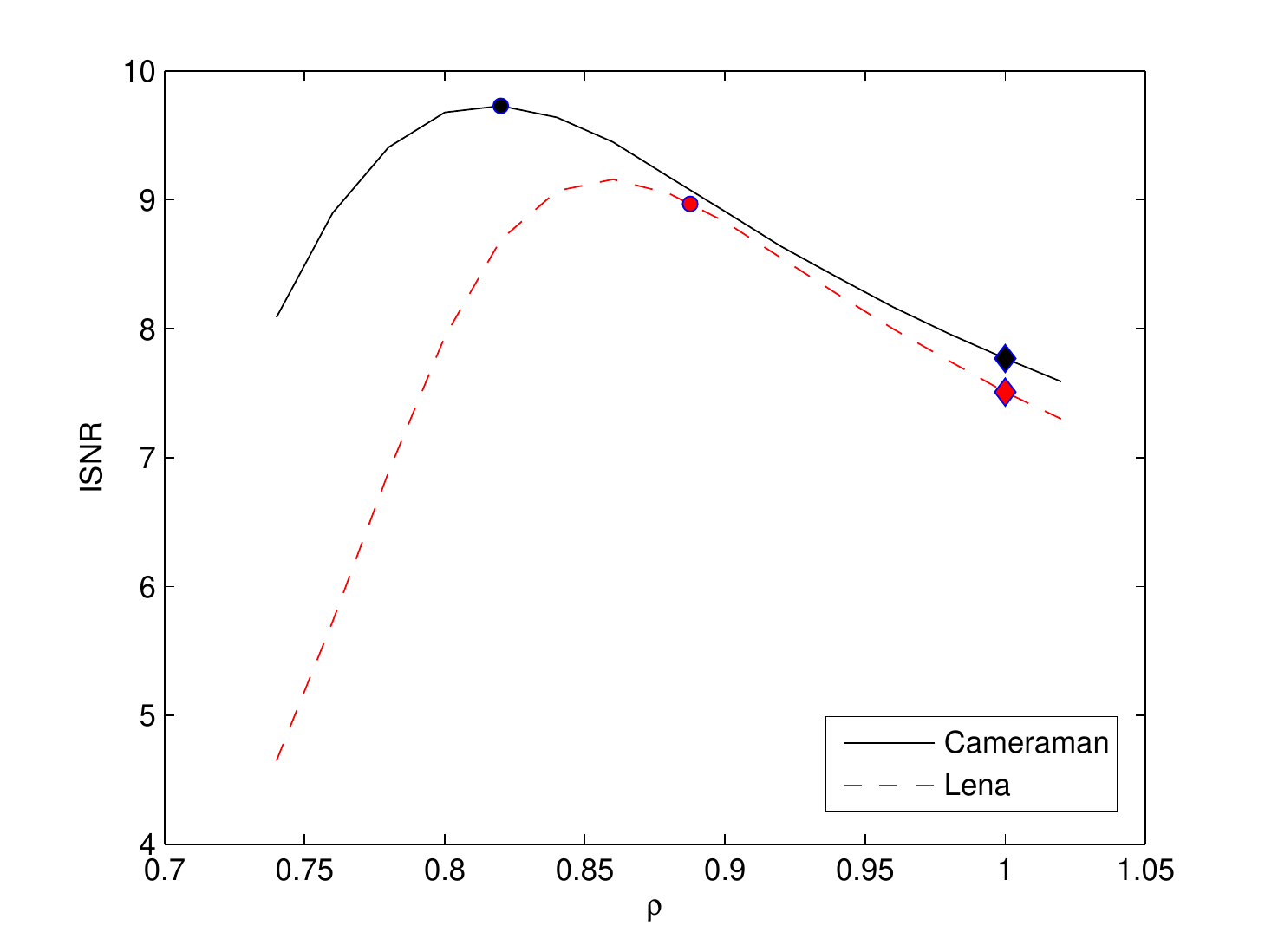}\\
\includegraphics[width=1\linewidth]{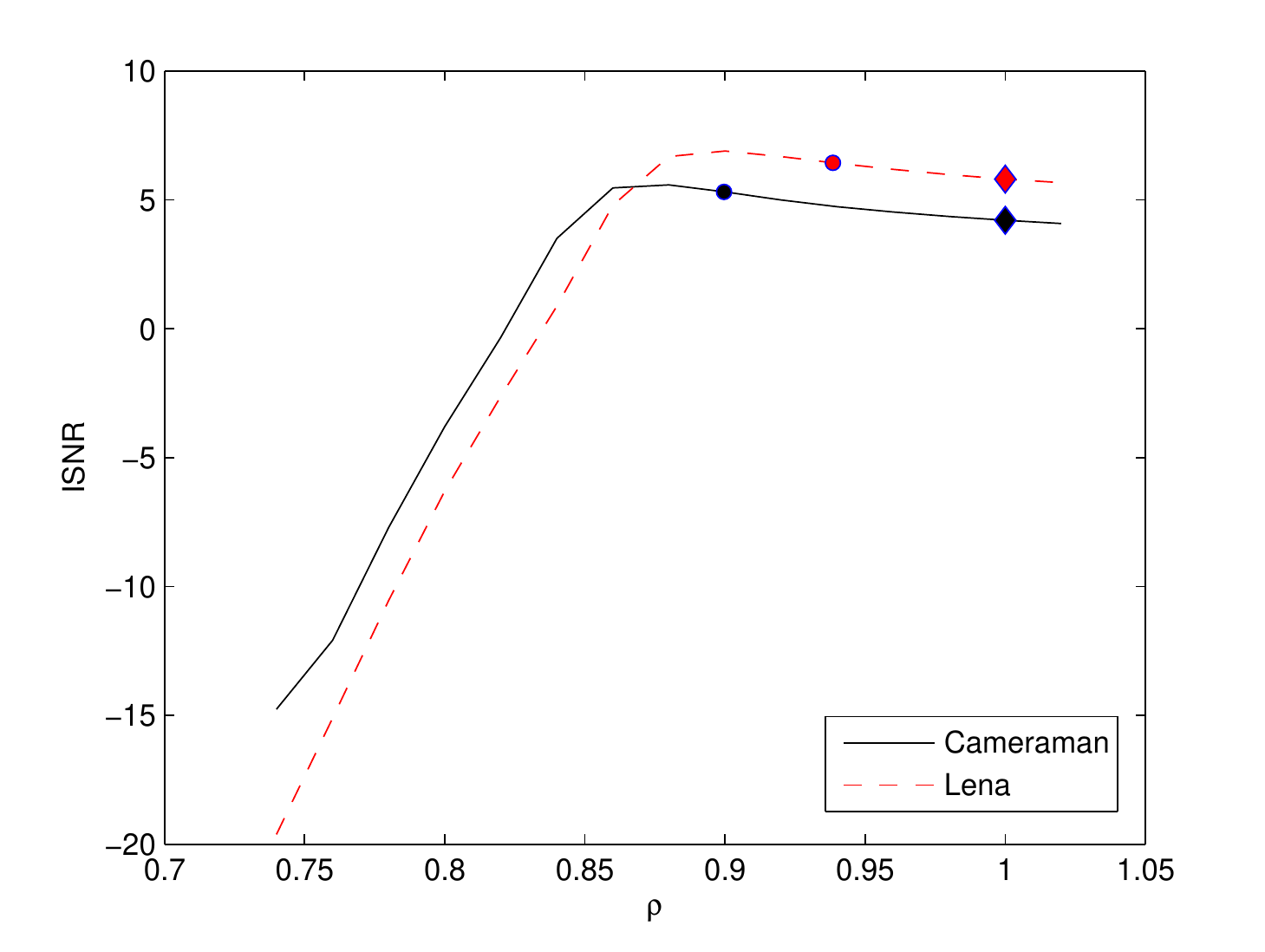}
\label{fig:side:c}
\end{minipage}
\caption{Parameter $\rho$ versus ISNR for $Cameraman$ and $Lena$ images. The images are blurred by a uniform blur of size $9 \times 9$(first row) and a Gaussian blur of size
$25 \times 25$ with variance 1.6 (second row). The ISNR values obtained by the proposed method and by  $\rho =1$ are marked by $"\diamondsuit"$ and $"\bullet"$, respectively.}
\end{figure*}


\subsection{Experiment 2 -- Comparison with the State-of-the-Art}

In this subsection, we compare the proposed algorithm with five state-of-the-art methods in standard test settings for image deconvolution.

Table I describes the different point spread functions (PSFs) and different variances of  white Gaussian additive noise. We remark that these benchmark experiment settings are commonly used in many previous works \cite{J.Portilla}\cite{Dong2}.
\begin{table}[h]
\caption{Experiment settings with different blur kernels and different values of noise variance $\sigma^{2}$ for pixel values in [0,255].}
  \begin{center}\hspace*{+0cm}
  \begin{tabular}{|c|c|c|}\hline
  Scenario & PSF & $\sigma^{2}$  \\
  \hline
  1 & $1/(1+i^{2}+j^{2})$, for $i,j=-7,...,7$ &2  \\
   \hline
  2 & $1/(1+i^{2}+j^{2})$, for $i,j=-7,...,7$ & 8  \\
   \hline
  3 &  $9\times 9$ uniform kernel (boxcar) & $\approx$ 0.3 \\
   \hline
  4 & $[1\ 4\ 6\ 4\ 1]^{T}[1\ 4\ 6\ 4\ 1]/256$ & 49 \\
   \hline
  5 &  $25\times 25$ Gaussian with std = 1.6 & 4 \\
   \hline
  \end{tabular}
  \end{center}
\end{table} \label{table1}

In this section, the proposed GFD algorithm is compared
with five currently state-of-the-art deconvolution algorithms, i.e.,
ForWaRD \cite{R.Neelamani}, APE-ADMM \cite{Chuan}, L0-ABS \cite{J.Portilla}, SURE-LET \cite{Xue}, and BM3DDEB \cite{K.Dabov}.
The default parameters by the authors are applied for
the developed algorithms. We emphasize that the APE-ADMM \cite{Chuan} is also an adaptive parameter selection method for total variance deconvolution. The comparison of our test results for different experiments against the other five methods in terms of ISNR are shown in Table II.

\begin{table*}[t]
\caption{Comparison of the output ISNR(dB) of the proposed deblurring algorithm. BSNR(Blurred Signal-to-noise ratio) is defined as $BSNR = 10\log_{10}Var(y)/N^{2}\sigma^{2}$, where $Var()$ is the variance.}
\centering
\begin{tabular}{|c|c|c|c|c|c|c|c|c|c|c|}
\hline
& \multicolumn{5}{c|}{Scenario} & \multicolumn{5}{|c|}{Scenario} \\
\cline{1-11}
&1 & 2 & 3 & 4&5 & 1 & 2 & 3 & 4&5\\

\hline
Method & \multicolumn{5}{c|}{Cameraman ($256 \times 256$)} & \multicolumn{5}{|c|}{House ($256 \times 256$)} \\

\hline
BSNR& 31.87  & 25.85& 40.00   & 18.53&   29.19    & 29.16 &  23.14 & 40.00   & 15.99&  26.61  \\
\hline
\hline
ForWaRD& 6.76  & 5.08 & 7.40  & 2.40 & 3.14 & 7.35 & 6.03 &9.56 &3.19 &3.85  \\
\hline
APE-ADMM&  7.41& 5.24 &8.56& 2.57 &3.36 & 7.98 &6.57 &10.39& 4.49& 4.72   \\
\hline
L0-Abs& 7.70 &5.55 &9.10& 2.93 &3.49  & 8.40 &7.12& 11.06& 4.55 &4.80 \\
\hline
SURE-LET& 7.54  &  5.22 & 7.84    &  2.67&   3.27    & 8.71 &  6.90 & 10.72    &  4.35&   4.26    \\
\hline
BM3DDEB& 8.19& 6.40 &8.34& 3.34 &3.73 & 9.32 &\textbf{8.14} &10.85& 5.13& 4.79\\
\hline
GFD& \textbf{8.38} & \textbf{6.52}& \textbf{9.73}  &  \textbf{3.57} & \textbf{4.02}  &\textbf{9.39}&7.75 & \textbf{12.02}    & \textbf{5.21}&   \textbf{5.39}  \\
\hline
\hline
& \multicolumn{5}{c|}{Scenario} & \multicolumn{5}{|c|}{Scenario} \\
\cline{1-11}
&1 & 2 & 3 & 4&5  & 1 & 2 & 3 & 4&5 \\

\hline
Method & \multicolumn{5}{c|}{Lena ($512 \times 512$)} & \multicolumn{5}{|c|}{Man ($512 \times 512$)} \\
\hline
BSNR& 29.89 & 23.87 & 40.00 &  16.47&  27.18  & 29.72  & 23.70 & 40.00  & 16.32&  27.02  \\
\hline
\hline
ForWaRD& 6.05& 4.90 &6.97 &2.93& 3.50 & 5.15 &3.87& 6.47&2.19& 2.71\\
\hline
APE-ADMM& 6.36  &  4.98 & 7.87    &  3.52&   3.61  & 5.82 &4.28& 7.14 &2.58& 2.98\\
\hline
L0-Abs& 6.66  &  5.71 & 7.79    &  4.09&   4.22  & 5.74& 4.02& 7.19& 2.61& 3.00  \\
\hline
SURE-LET& 7.71  &  5.88 & 7.96   &  4.42&   4.25    & 6.01  &  4.32 & 6.89    &  2.75&   3.01  \\
\hline
BM3DDEB&7.95 &  6.53 & 8.06    &  \textbf{4.81}&   4.37    & \textbf{6.34} &4.81&6.99&3.05 &3.22 \\
\hline
GFD& \textbf{8.12}  & \textbf{6.65}& \textbf{8.97}   & 4.77&   \textbf{4.95} &6.29 &  \textbf{4.83} & \textbf{7.67}   &  \textbf{3.11}&   \textbf{3.50}   \\
\hline
\end{tabular}
\end{table*}\label{table2}

 From Table II, we can see that our GFD algorithm achieves highly competitive performance and outperforms the other leading
deconvolution algorithms in the most cases. The highest ISNR results in
the experiments are labeled in bold.

\begin{figure*}[ht]
  \centering
  \centerline{\includegraphics[width=1\linewidth]{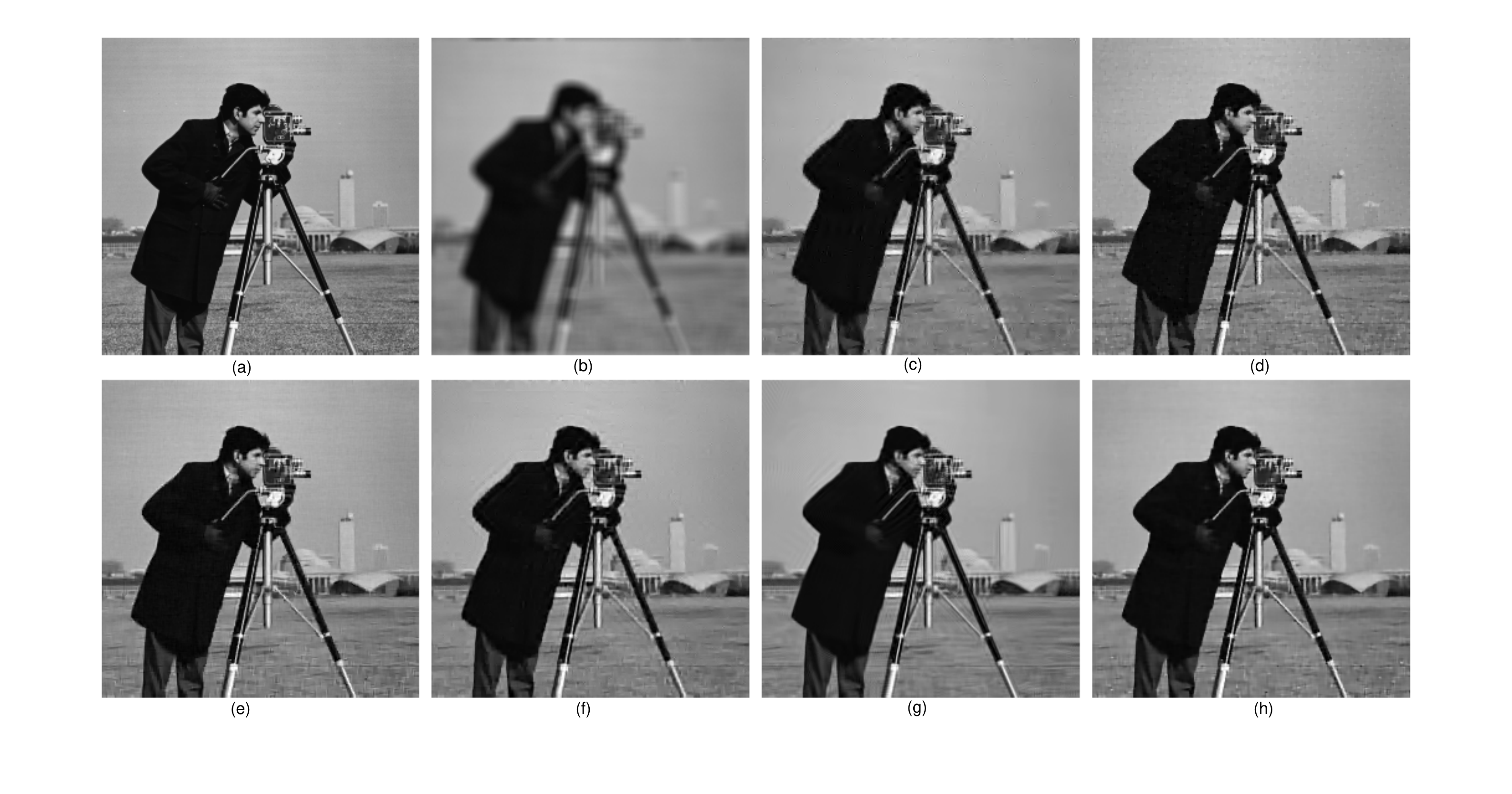}}
 \vspace{-1.0cm}
\caption{Visual quality comparison of image deblurring on gray image
$Cameraman$ (256$\times$256). (a) original image, (b) noisy and blurred image (scenario 3), (c) the result of ForWaRD  (ISNR=7.40dB), (d) the result of APE-ADMM  (ISNR=8.56dB), (e) the result of L0-ABS  (ISNR=9.10dB), (f) the result of SURE-LET (ISNR=7.84dB), (g)BM3DDEB result (ISNR = 8.34dB ), and (h) the result of our method (ISNR = 9.73dB).}
\label{p.2}
\end{figure*}

One can see that ForWaRD method is less competitive for cartoon-like images than for these less structured images.
The TV model is substantially
outperformed by our method for complicated images like $Cameraman$ and $Man$ with lots of disordered features and irregular edges, though it is well-known for its
outstanding performance on regularly-structured images such
as $Lena$ and $House$.

SURE-LET and L0-ABS achieve higher average ISNR than ForWaRD and APE-ADMM,
while our algorithm outperforms SURE-LET  by 1.25 dB for
scenario 3 and outperforms L0-ABS  by 0.84 dB for the scenario 2, respectively.
But L0-ABS cannot obtain top performance with all kinds of images and degradations without
suitable for the image sparse representation and the mode parameters to the observed image.
SURE-LET approach is applicable for periodic boundary conditions, and can
be used in various practical scenarios. But it also loses some details in the restored images.

BM3DDEB, which achieves the best performance on average,
One can found that our method and BM3DDEB produce similar results, and achieve
significant ISNR improvements over other leading
methods. In average, our algorithm outperforms BM3DDEB
by (0.095dB, -0.0325dB, 1.04dB,  0.0825dB, 0.46dB) for the five settings, respectively. In Figures 4$\sim$7, we show the visual comparisons of the deconvolution algorithms, from which it can be observed
that the GFD approach produces sharper and cleaner image edges than other five methods.

\begin{figure*}[ht]
  \centering
  \centerline{\includegraphics[width=1\linewidth]{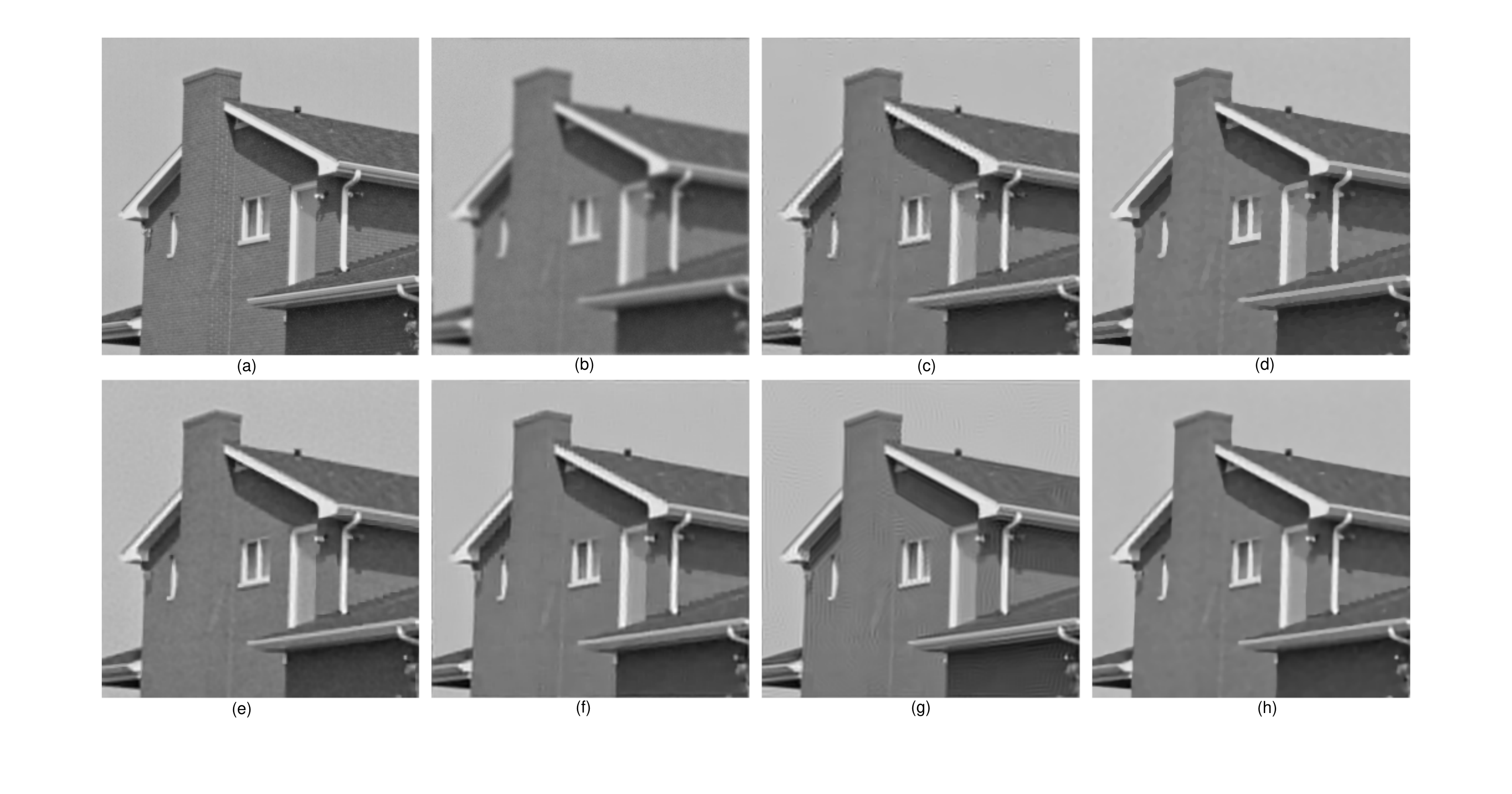}}
 \vspace{-1.0cm}
\caption{Visual quality comparison of image deblurring on gray image
$House$ (256$\times$256). (a) original image, (b) noisy and blurred image (scenario 5), (c) the result of ForWaRD  (ISNR=3.85dB), (d) the result of APE-ADMM  (ISNR=4.72dB), (e) the result of L0-ABS  (ISNR=4.80dB), (f) the result of SURE-LET (ISNR=4.26dB), (g)BM3DDEB result (ISNR = 4.79dB ), and (h) the result of our method (ISNR = 5.39dB).}
\label{p.3}
\end{figure*}

From Figure 4, one can evaluate the visual quality of some restored images. It
can be observed that our method is able to provides sharper
image edges and suppress the ringing artifacts better than BM3DDEB.
For $House$ image (Figure 5), the differences between the various methods are clearly visible: our algorithm introduces fewer artifacts than the other methods. In Figure 6, our method achieves good preservation of regularly-sharp edges and uniform
areas, while for $Man$ image (Figure 7), it preserves the finer details of the irregularly-sharp edges.

\begin{figure*}[!t]
  \centering
  \centerline{\includegraphics[width=1\linewidth]{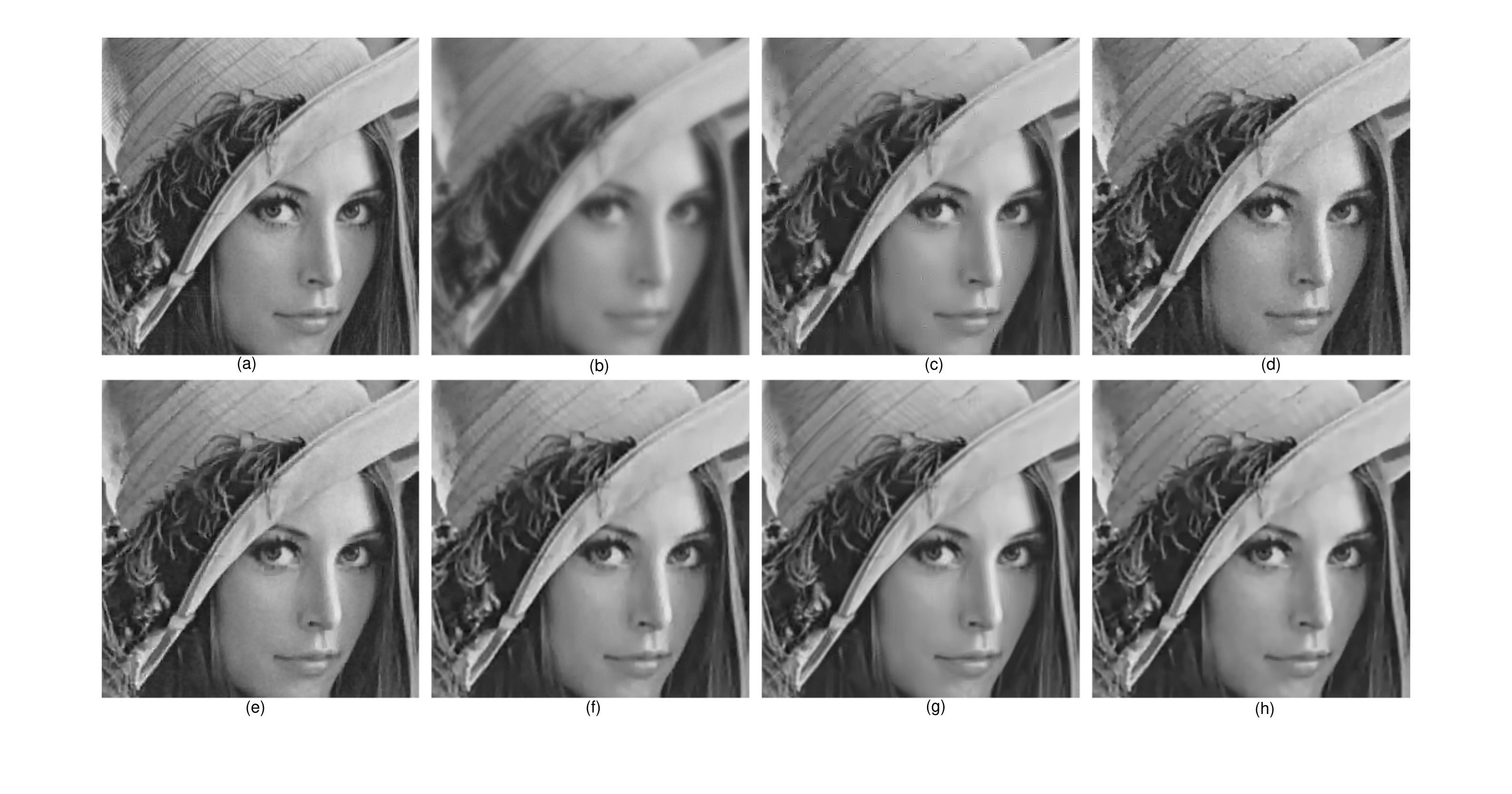}}
\vspace{-1.0cm}
\caption{Details of the image deconvolution experiment on image $Lena$ (512$\times$512). (a) original image, (b) noisy and blurred image (scenario 1), (c) the result of ForWaRD  (ISNR=6.05dB), (d) the result of APE-ADMM  (ISNR=6.36dB), (e) the result of L0-ABS  (ISNR=6.66dB), (f) the result of SURE-LET (ISNR=7.71dB), (g)BM3DDEB result (ISNR = 7.95dB ), and (h) the result of our method (ISNR = 8.12dB).}
\label{p.4}
\end{figure*}

\begin{figure*}[!t]
  \centering
  \centerline{\includegraphics[width=1\linewidth]{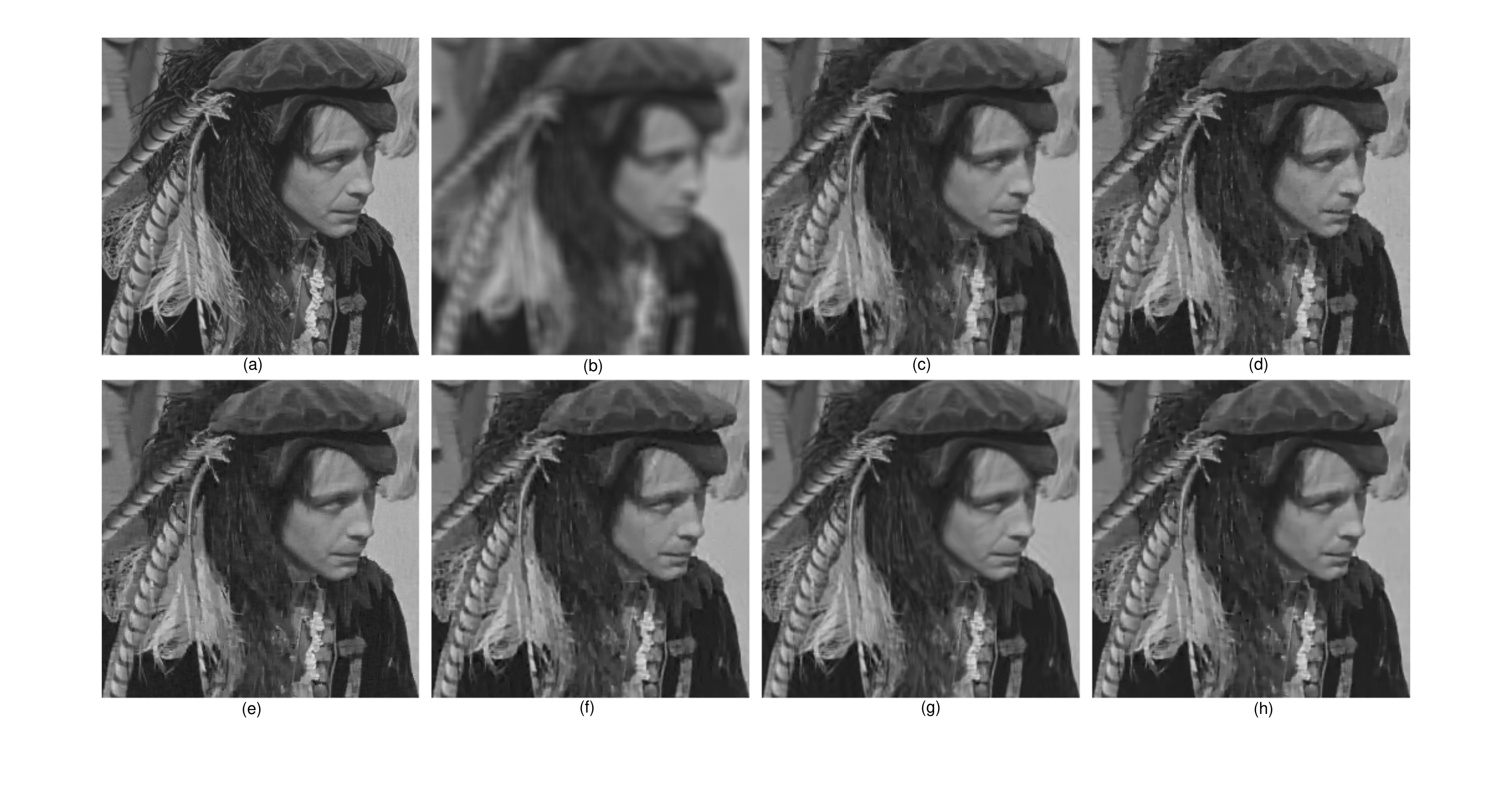}}
 \vspace{-1.0cm}
\caption{Details of the image deconvolution experiment on image $Man$ (512$\times$512). (a) original image, (b) noisy and blurred image (scenario 3), (c) the result of ForWaRD  (ISNR=6.47dB), (d) the result of APE-ADMM  (ISNR=7.14dB), (e) the result of L0-ABS  (ISNR=7.19dB), (f) the result of SURE-LET (ISNR=6.89dB), (g)BM3DDEB result (ISNR = 6.99dB ), and (h) the result of our method (ISNR = 7.67dB).}
\label{p.5}
\end{figure*}

\begin{figure*}[!t]
\begin{minipage}[t]{0.22\linewidth}
\centering
\includegraphics[width=1\linewidth]{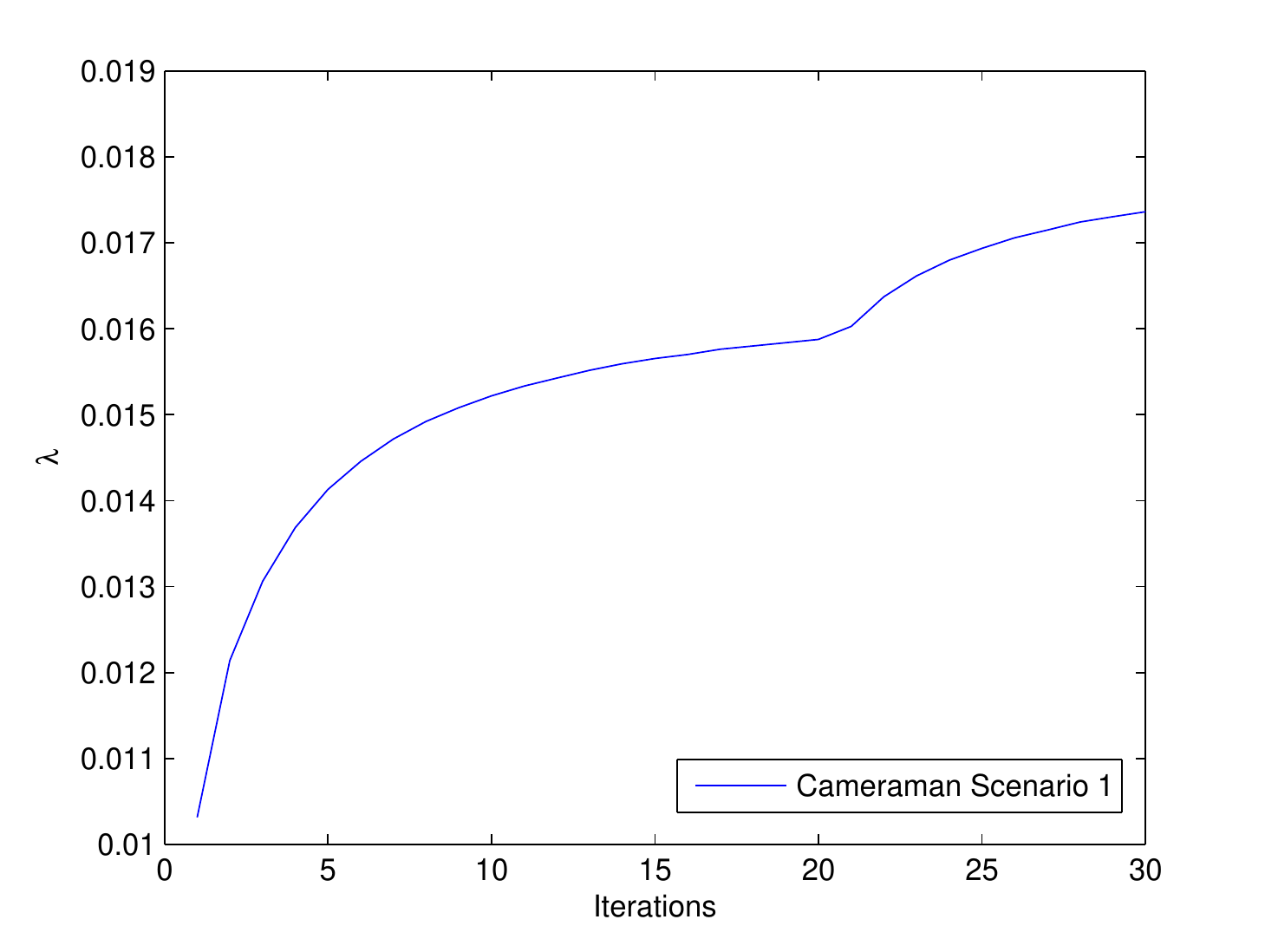}\\
\includegraphics[width=1\linewidth]{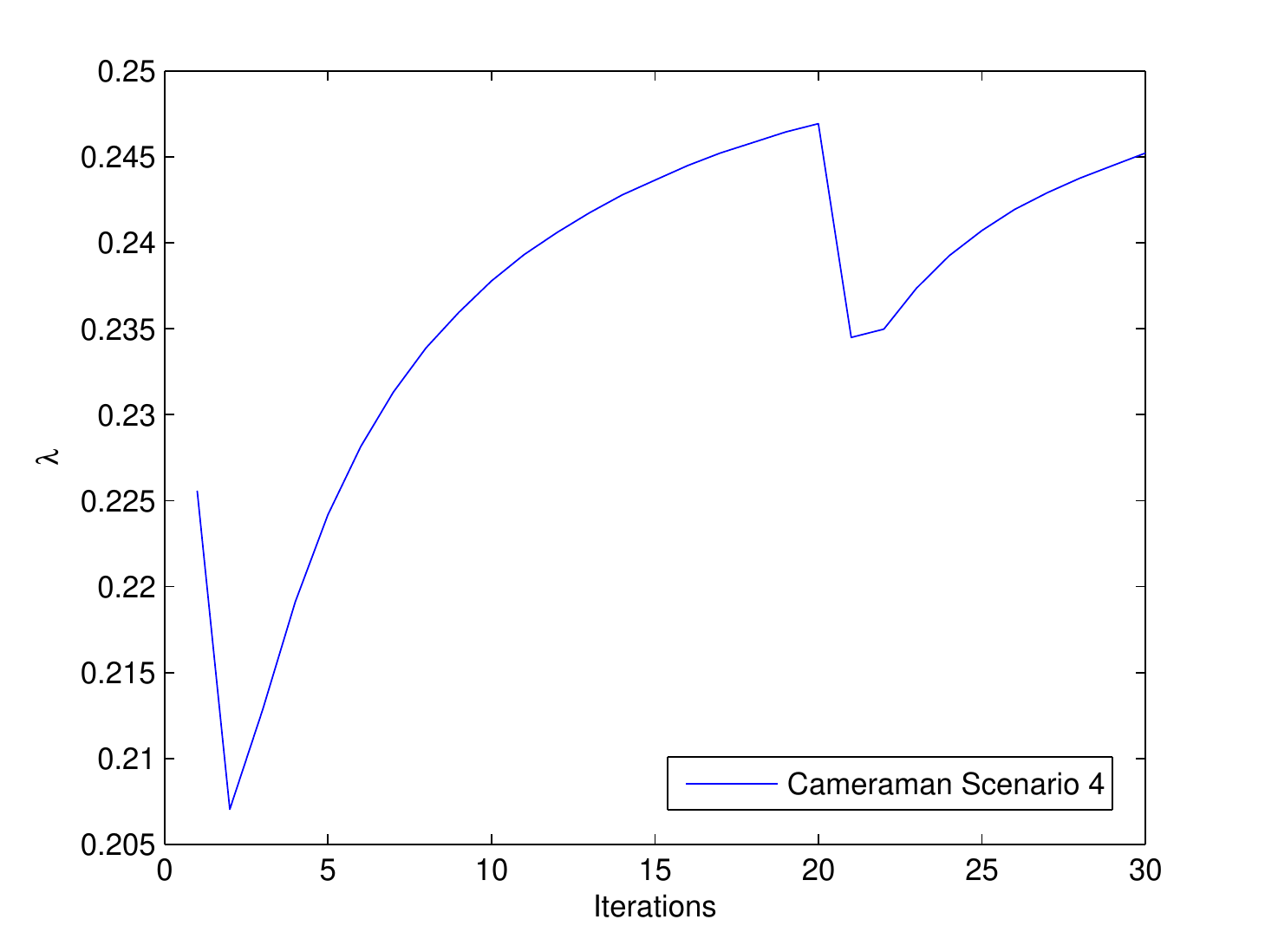}
\label{fig:side:a}
\end{minipage}%
\begin{minipage}[t]{0.22\linewidth}
\centering
\includegraphics[width=1\linewidth]{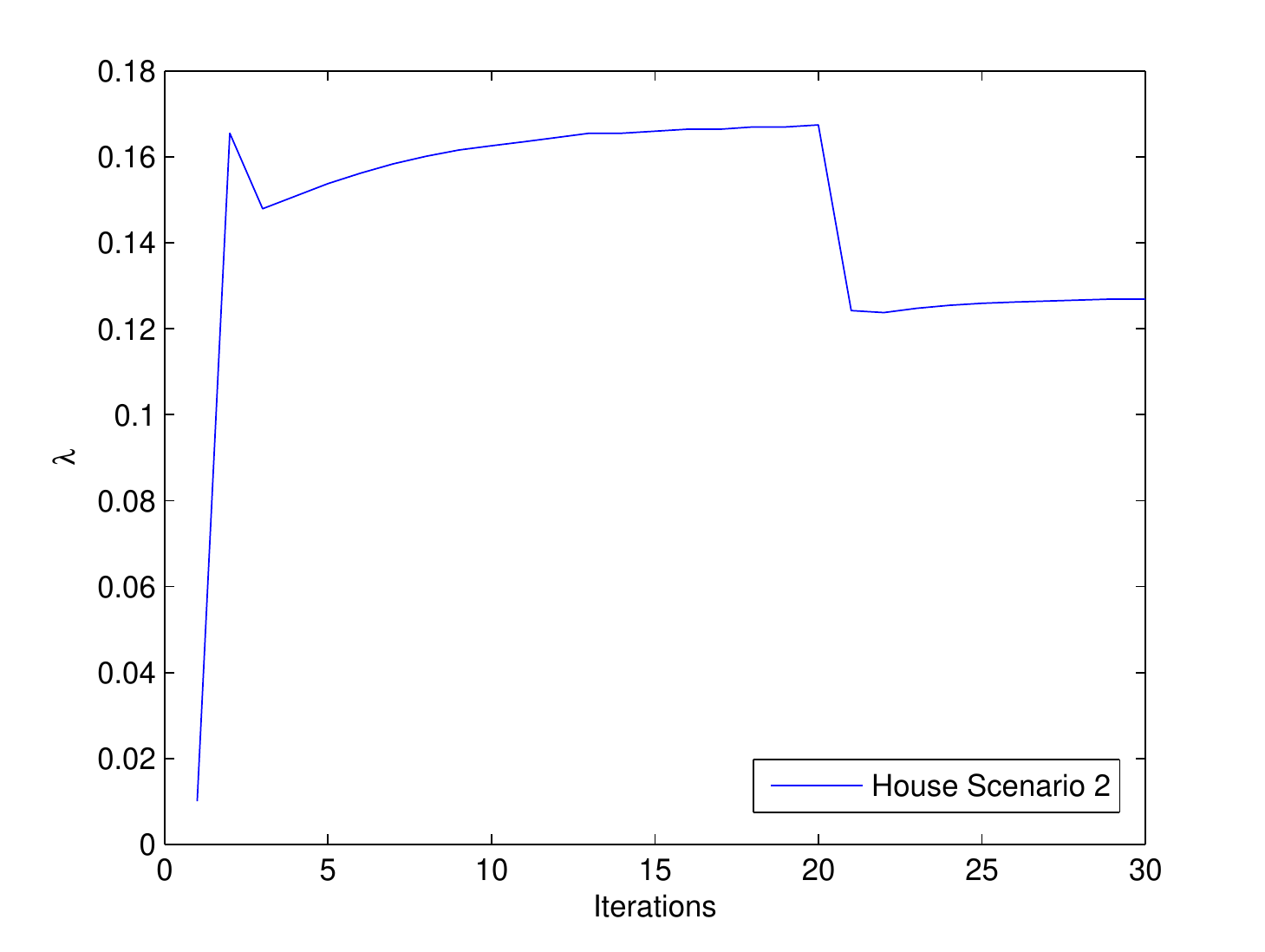}\\
\includegraphics[width=1\linewidth]{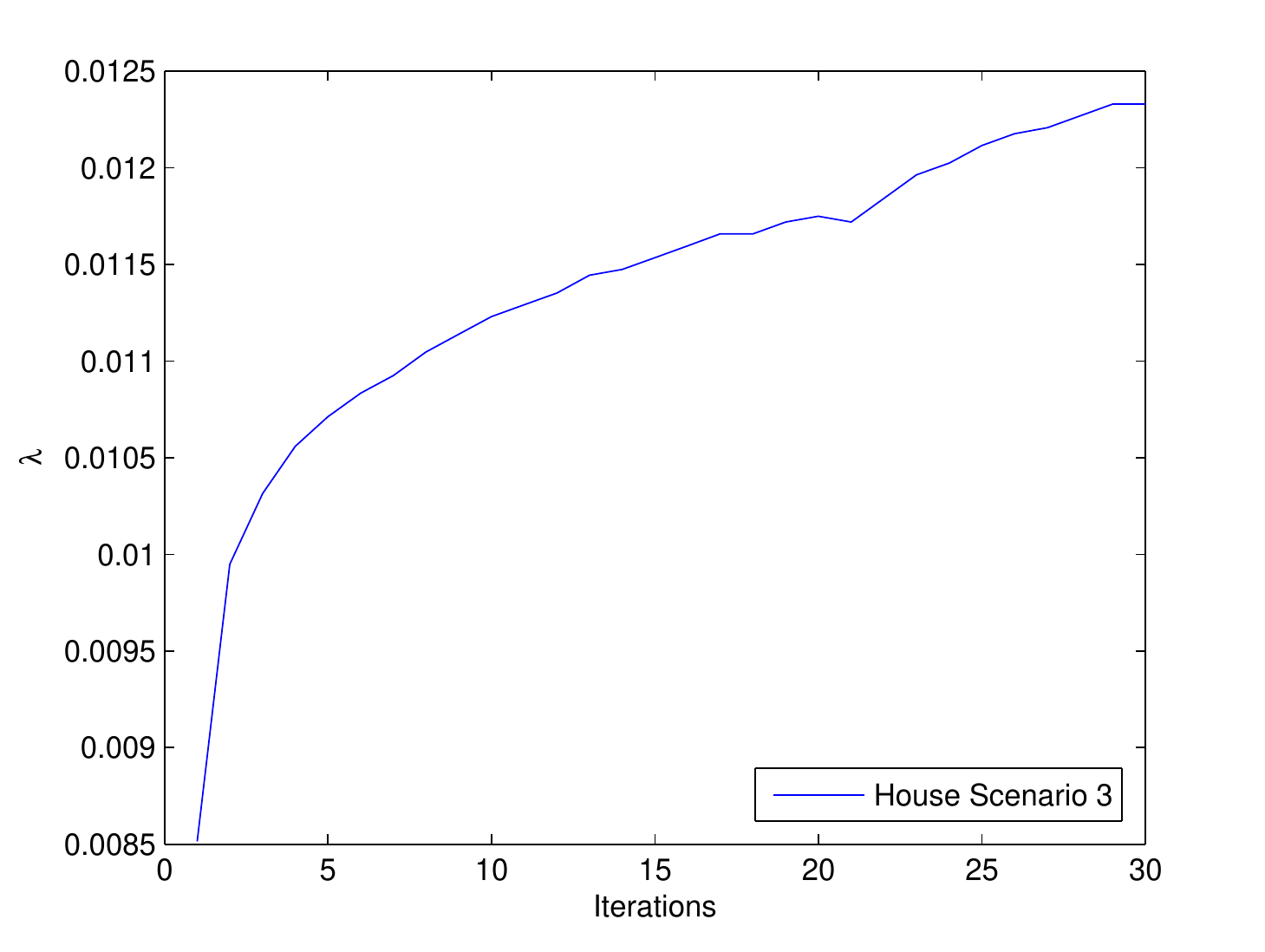}
\label{fig:side:b}
\end{minipage}
\begin{minipage}[t]{0.22\linewidth}
\centering
\includegraphics[width=1\linewidth]{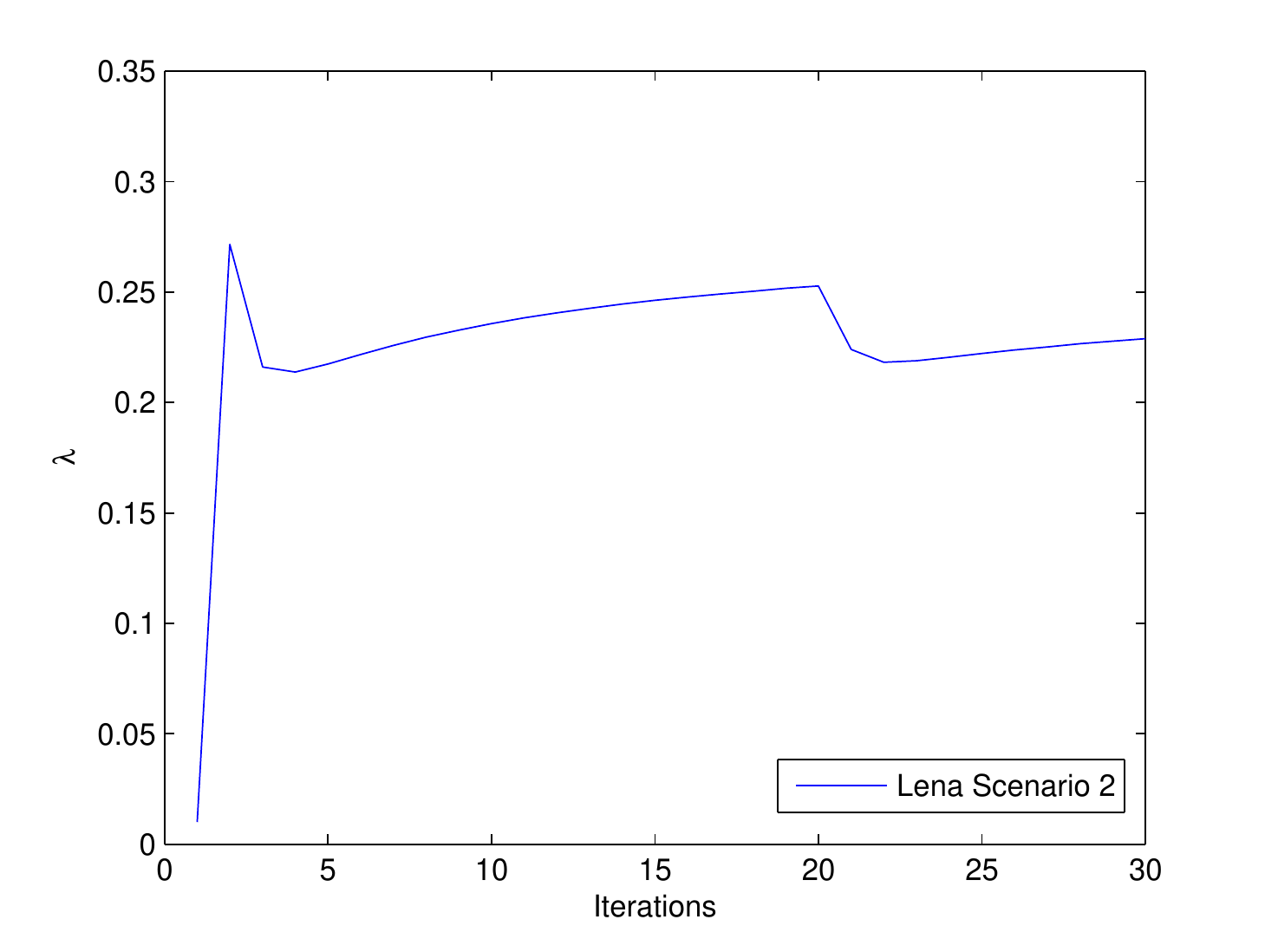}\\
\includegraphics[width=1\linewidth]{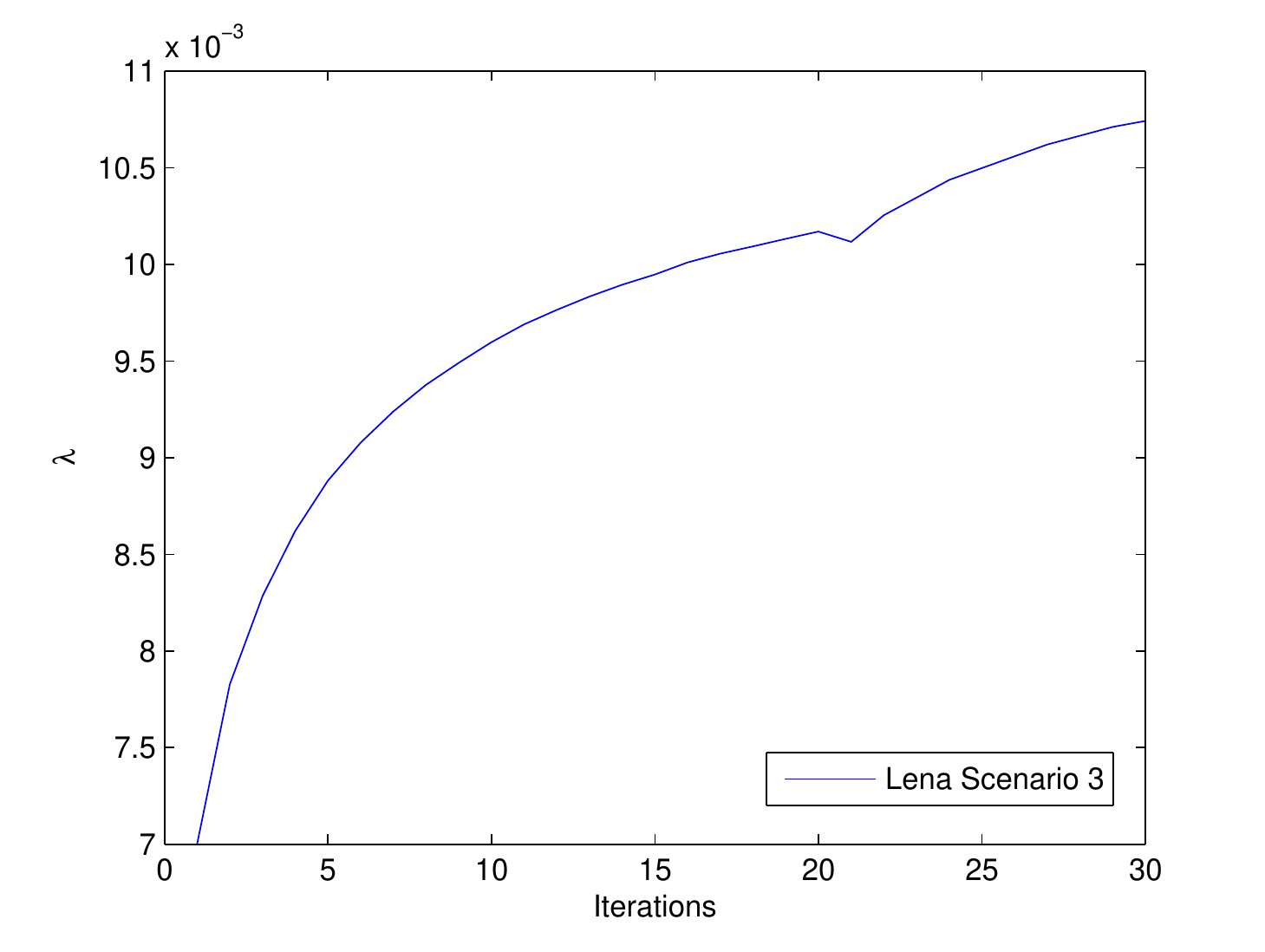}
\label{fig:side:c}
\end{minipage}
\begin{minipage}[t]{0.22\linewidth}
\centering
\includegraphics[width=1\linewidth]{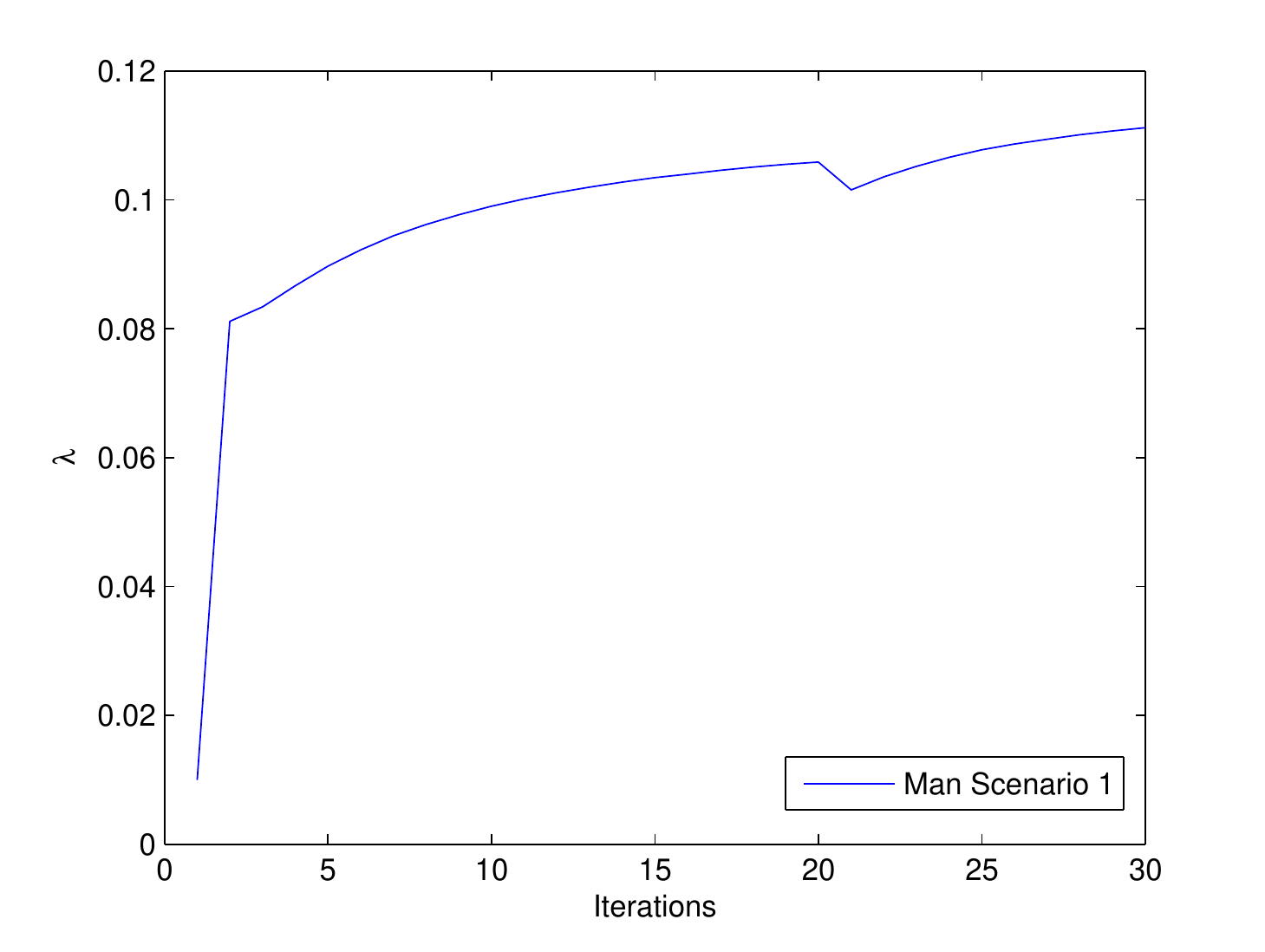}\\
\includegraphics[width=1\linewidth]{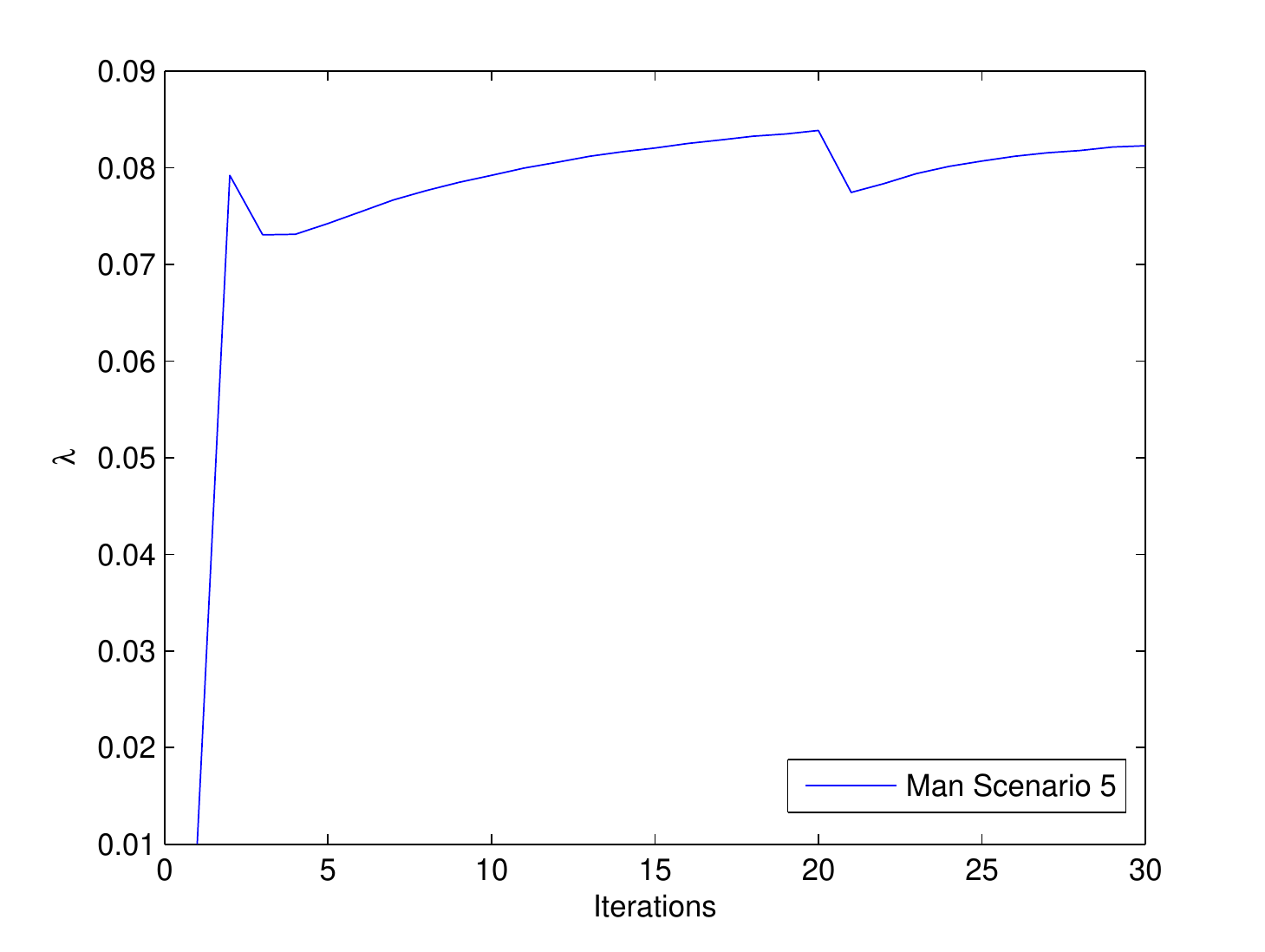}
\label{fig:side:d}
\end{minipage}
\caption{Regularization parameter versus iterations number. The $Cameraman$ image is tested by Scenario 1 and 4; The $House$ image is tested by Scenario 2 and 3; The $Lena$ image is tested by Scenario 2 and 3; The $Man$ image is tested by Scenario 1 and 5.}
\end{figure*}

In Figure 8, we plotted a few curves of different $\lambda$ values versus iteration number, which were obtained
from scenario 1 and 4 using $Cameraman$ image, scenario 2 and 3 using $House$ image, scenario 2 and 3 using $Lena$ image, scenario 1 and 5 using $Man$ image respectively.
Hence, we emphasize that our approach automatically
choose the regularization parameter at each iteration, unlike some of the other deconvolution
algorithms such as \cite{Y.Wang}, one has to manually tune the parameters, e.g., by running the method many times
to choose the best one by error and trial.

\subsection{Experiment 3 -- Convergence}

Since the guided image filter is highly nonlinear, it is difficult to prove the global convergence of our algorithm in theory.
In this section, we only provide empirical evidence to show the stability of the proposed deconvolution algorithm.

\begin{figure}[!t]
\begin{minipage}[t]{0.45\linewidth}
\centering
\includegraphics[width=1\linewidth]{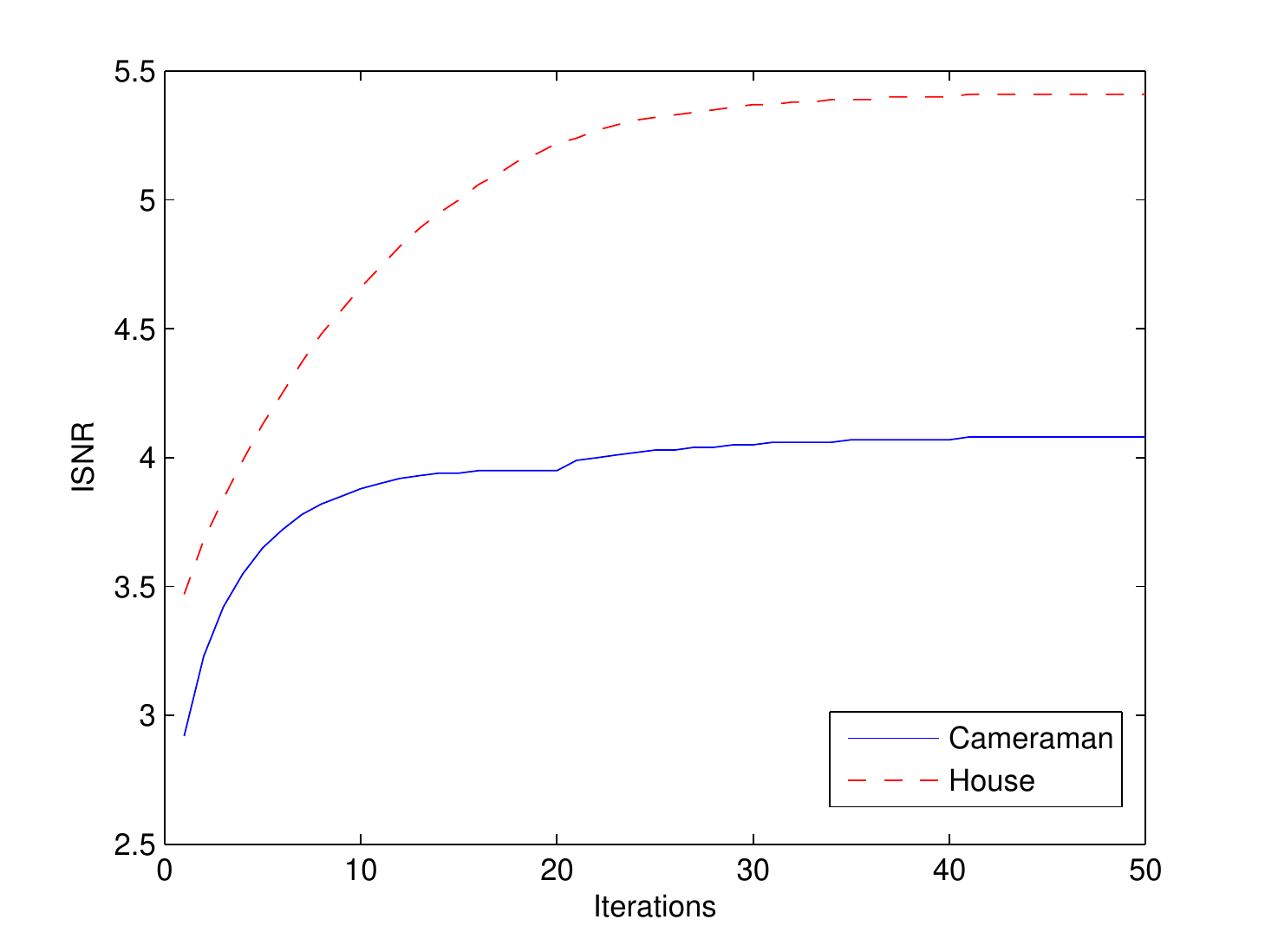}\\
\label{fig:side:a}
\end{minipage}%
\begin{minipage}[t]{0.45\linewidth}
\centering
\includegraphics[width=1\linewidth]{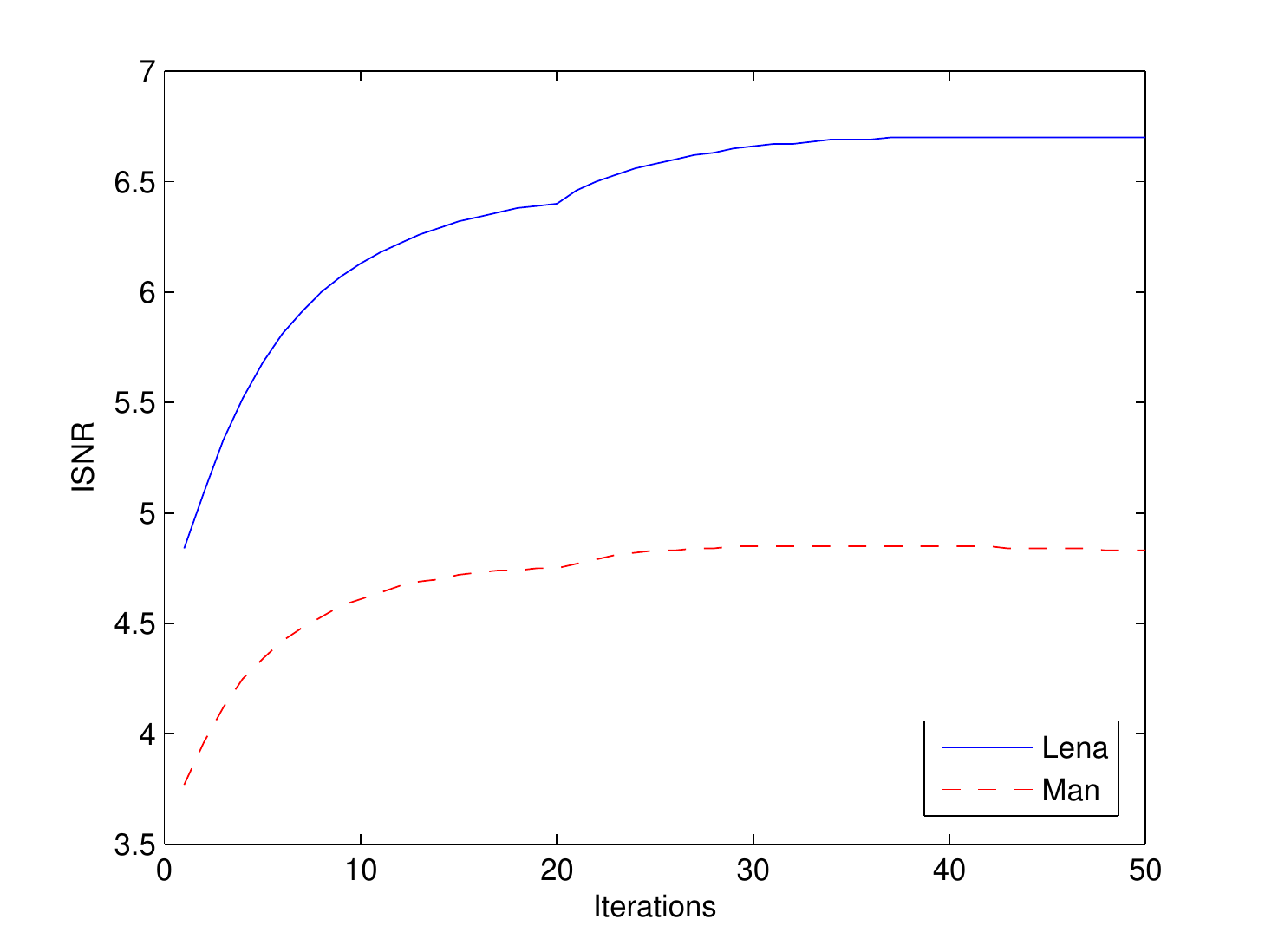}\\
\label{fig:side:b}
\end{minipage}
\caption{Change of the ISNR with iterations for the different settings of the
proposed algorithm. Left: deblurring of $Cameraman$ and $House$ images, scenario 5; Right: deblurring of $Lena$ and $Man$ images, scenario 2. }
\label{p.3.3}
\end{figure}

In Figure \ref{p.3.3}, we show the convergence properties of the GFD algorithm
for test images in the cases of scenario 5 (blur kernel is $25\times 25$ Gaussian with std = 1.6, $\sigma = 2$) and scenario
2 (PSF = $1/(1+i^{2}+j^{2})$, for $i,j=-7,...,7$, $\sigma = \sqrt{8}$) for four test
images. One can see that all the ISNR curves grow monotonically with the increasing of
iteration number, and finally become stable and flat. One can also found that 30 iterations are typically sufficient.

\subsection{Analysis of Computational Complexity}

In this work, all the experiments are performed in Matlab R2010b
on a PC with Intel Core (TM) i5 CPU processor (3.30 GHz), 8.0G memory, and
Windows 7 operating system.

In MATLAB simulation, we have obtained times per iteration of 0.057 seconds using $256 \times 256$ image, and about 30 iterations are enough.

The most computationally-intensive part of our
algorithm is guided image filtering (3 times in one iteration).  
We mention that the guided filter can be simply sped up with subsampling, and this will lead to a speedup of $10$ times with almost no visible degradation\cite{He3}.

\section{Conclusion}
We have developed an image deconvolution algorithm based on  an  explicit image filter : guided filter, which has been proved to be more effective than the bilateral filter in several applications.  We first introduce guided filter into
the image restoration problem and propose an efficient method, which obtains high quality results.
The simulation results show that our method outperforms some existing competitive deconvolution method.
We find remarkable how such a simple method compares to other much more sophisticated methods.
Based on Morozov's discrepancy principle, we also propose a simple but effective method to automatically determine the regularization parameter at each iteration.


%

\appendices
%
%
%
%
%

\ifCLASSOPTIONcaptionsoff
  \newpage
\fi

\end{document}